\documentclass[a4paper,fleqn]{cas-sc}

\usepackage[authoryear]{natbib}

\def\tsc#1{\csdef{#1}{\textsc{\lowercase{#1}}\xspace}}
\tsc{WGM}
\tsc{QE}
\tsc{EP}
\tsc{PMS}
\tsc{BEC}
\tsc{DE}

\usepackage{xcolor}
\colorlet{baseColor}{white}

\usepackage{tabularx}
\usepackage{booktabs}
\usepackage{enumitem}
\usepackage{comment}
\usepackage{lipsum}
\usepackage{amsmath}
\usepackage{url}
\usepackage{graphicx}
\usepackage{capt-of}

\usepackage{tikz}
\usetikzlibrary{arrows.meta,positioning}

\newcommand{\human}{\textsc{Human}}
\newcommand{\freellm}{\textsc{Free-LLM}}
\newcommand{\arb}{\textsc{ARB}}
\newcommand{\arbfull}{\arb{}: Authorship-Rewriting Benchmark}
\newcommand{\htol}{\textsc{H2L}}
\newcommand{\llmtol}{\textsc{LLM2L}}
\newcommand{\tprfpr}{\mbox{TPR@1\%FPR}}

\usepackage{hyperref}

\providecommand{\DOIprefix}{}
\renewcommand{\DOIprefix}{}

\providecommand{\doi}[1]{}
\renewcommand{\doi}[1]{\url{https://doi.org/#1}}
\providecommand{\URLprefix}{}
\renewcommand{\URLprefix}{}

\begin{document}
\let\WriteBookmarks\relax
\def\floatpagepagefraction{1}
\def\textpagefraction{.001}

\shorttitle{ARB-Dataset}

\title[mode = title]{ARB: A Matched Authorship-Rewriting Benchmark Dataset for AI-Text Detector Evaluation}

\author[1]{Gaetano Perrone}[type=author,
orcid=0000-0001-7511-2910]

\cormark[1]

\ead{gaetano.perrone@unina.it}

\affiliation[1]{organization={Department of Electrical Engineering and Information Technology, University of Napoli Federico II},
    addressline={Via Claudio 21}, 
    city={Naples},
postcode={80125}, 
country={Italy}}

\author[1]{Simon Pietro Romano}[bioid=1]

\cortext[cor1]{Corresponding author}

\begin{abstract}
Standard AI-text detection benchmarks compare human-written text against text generated directly by large language models (LLMs). While prior work has shown that rewriting and paraphrasing can degrade detector performance, it remains unclear whether performance measured on this conventional benchmark predicts detector behavior when human-authored content is rewritten by an LLM. To address this gap, we introduce \arbfull{}, built from 1,800 human source texts (600 each from XSum, WritingPrompts, and OpenWebText) and four open-weight generators (Llama-3.2-3B, Qwen2.5-7B, Mistral-7B, Gemma-2-9B). Each source item yields four matched variants: human-written (\human{}), direct LLM generation (\freellm{}), LLM-rewritten human text (\htol{}), and same-generator LLM-rewritten LLM text (\llmtol{}). We evaluated five detectors (FastDetectGPT, Binoculars-falcon-7b, RADAR, BERT-Defense, RoBERTa-Defense) at a strict 1\%-false-positive operating point (\tprfpr{}). FastDetectGPT and Binoculars-falcon-7b detected 91.2\% and 93.5\% of direct LLM text, but only 30.8\% and 15.1\% of human text an LLM had rewritten, a drop of 60--78 percentage points. The same detectors retained 78.3\% and 83.0\% recall when LLM text was rewritten by the same model, a much smaller decline of 10--13 points. RADAR followed the same pattern (66.8\% to 12.2\%), while BERT-Defense and RoBERTa-Defense stayed below 3\% recall across all regimes. These results show that detector performance measured on the conventional human-vs-LLM benchmark does not transfer to human-authored text revised by an LLM, even though the same detectors remain largely robust to LLM-only rewriting.

\end{abstract}

\begin{highlights}
\item New four-regime benchmark separates AI-authorship from AI-mediated rewriting
\item Top detectors reach 91-94\% recall on direct AI-generated text
\item Detector recall falls to 15-31\% when human text is AI-rewritten.
\item Rewriting AI text with the same model keeps recall near 78-83\%.
\item Standard human-vs-AI benchmarks overestimate robustness to AI rewriting.
\end{highlights}


\maketitle

\section{Introduction}
\label{sec:introduction}

Large language models (LLMs) are nowadays used in a broad range of writing workflows, including drafting, rewriting, summarization, polishing, and style transfer. Their downstream use is not limited to free-form generation but also includes assisted composition and the transformation of existing text~\citep{10.1145/3649506}. AI-text detection, however, is still often evaluated as a binary problem: distinguishing human-written text from text generated directly by a model~\citep{gehrmann2019gltr,ippolito2020automatic,mitchell2023detectgpt,li2024mage}. This standard benchmark is a necessary baseline, but it is incomplete as an empirical test of detector robustness. This paper therefore addresses a benchmark-validity question rather than only a paraphrase-robustness question.

The core limitation is that standard \human{} vs. LLM benchmarks conflate two factors. The first is \emph{content origin}: whether the ideas, facts, discourse structure, and semantic content originate from a human author or from an LLM. The second is \emph{linguistic surface}: whether the final wording is human-written, freely generated by an LLM, or mediated by an LLM through rewriting. These factors can diverge in realistic workflows. A student, journalist, analyst, or software engineer may write an initial draft and then use an LLM to improve fluency. In that case, the final text has human-origin content but an LLM-mediated surface. Conversely, an LLM-generated text may be passed again through the same LLM while remaining LLM-origin. A detector score may therefore reflect direct machine authorship, machine-mediated rewriting, domain artifacts, decoding artifacts, generator-specific cues, or interactions among these signals.

Paraphrase-oriented benchmarks have shown that rewriting, humanization, and adversarial transformation can substantially degrade detector performance~\citep{krishna2023paraphrasing,sadasivan2023reliably,10179387,shi-etal-2024-red,masrour2025damage}. Benchmarks that include human paraphrases, LLM paraphrases, or mixed human--machine text confirm this pattern at scale~\citep{lau2025effects,wang2024m4gtbench,wu2024detectrl,zha2025padben}. Large-scale evaluations further show that detector performance varies with generator family, domain, language, and attack type~\citep{wang2023m4,dugan2024raid,li2024mage,ayoobi2025shield,stowe2026spotlights}. These works establish that paraphrasing and rewriting can break detectors, but they primarily evaluate detector degradation under transformed text. They do not directly test whether performance measured on the conventional \human{} vs.\ \freellm{} benchmark transfers uniformly across matched authorship--surface regimes.

The main empirical contribution is therefore not another demonstration that rewriting degrades detectors, but evidence that performance measured under the conventional \human{} vs.\ \freellm{} benchmark does not transfer uniformly to matched LLM-mediated rewriting regimes.

\arbfull{} operationally contrasts content-origin and LLM-mediated surface regimes under a matched benchmark-transfer design. It does not estimate a pure causal effect of source origin. Each matched source item anchors a \human{} reference, a direct \freellm{} generation, a human-origin LLM-mediated rewrite (\htol{}), and a same-generator second-pass rewrite of the corresponding LLM output (\llmtol{}). This design turns rewriting from a generic attack condition into a diagnostic comparison: if \htol{} degrades while \llmtol{} remains close to \freellm{}, then the conventional direct-generation benchmark is not a reliable proxy for human-origin LLM-mediated writing. We evaluate performance primarily at a conservative low-false-positive operating point, with global ranking separability as a secondary view, since false positives on human-authored or human-origin text can be costly in educational, scientific, and organizational settings~\citep{LIANG2023100779,website-new-ai-classifier-for-indicating-ai-written-text}.

The remainder of the paper is organized as follows. Section~\ref{sec:research_objectives} states the research objectives and contributions. Section~\ref{sec:related_work} reviews prior work on AI-text detection, rewriting robustness, benchmark confounding, and low-false-positive evaluation, and situates the objectives relative to the closest existing benchmarks. Section~\ref{sec:benchmark_design} describes the matched four-regime benchmark design. Section~\ref{sec:evaluation_protocol} defines the evaluated detector families, metrics, block-level estimation procedure, paired deltas, and uncertainty analysis. Section~\ref{sec:results} reports the empirical results across regimes, detectors, datasets, and generators, addressing the research objectives directly. Section~\ref{sec:discussion} discusses the implications for detector robustness and benchmark design, including threats to validity and ethical considerations. The paper then concludes with the main findings and recommendations.

\section{Research Objectives}
\label{sec:research_objectives}

The basic objective of this work is to test whether detector performance measured under the conventional \human{} vs.\ \freellm{} benchmark is a valid proxy for detector performance under LLM-mediated rewriting, and to do so with a design that separates \emph{content origin} (human- or LLM-authored) from \emph{linguistic surface} (direct generation or LLM-mediated rewriting) rather than conflating them into a single ``rewritten'' class. The objectives are stated as follows:

\begin{itemize}[leftmargin=*]
    \item Determine whether performance estimated under direct-generation (\human{} vs.\ \freellm{}) benchmarking transfers to matched LLM-mediated rewriting regimes, or whether it overestimates robustness.
    \item Design a matched four-regime benchmark, anchored to shared source items within dataset--generator blocks, that isolates human-origin LLM-mediated rewriting (\htol{}) from same-generator LLM-origin second-pass rewriting (\llmtol{}).
    \item Quantify the operational \htol{}--\llmtol{} gap at a fixed, conservative low-false-positive operating point, alongside global ranking separability, and establish whether the gap is attributable to source origin, transformation strength, or both.
    \item Characterize how detector robustness varies across detector families, dataset domains, and generator models, so that conclusions are reported as stratified, block-level estimates rather than single aggregate scores.
\end{itemize}

\arbfull{} is the benchmark developed to meet these objectives. It operationally contrasts content-origin and LLM-mediated surface regimes under a matched benchmark-transfer design; it does not estimate a pure causal effect of source origin. Each matched source item anchors a \human{} reference, a direct \freellm{} generation, a human-origin LLM-mediated rewrite (\htol{}), and a same-generator second-pass rewrite of the corresponding LLM output (\llmtol{}). This design turns rewriting from a generic attack condition into a diagnostic comparison: if \htol{} degrades while \llmtol{} remains close to \freellm{}, then the conventional direct-generation benchmark is not a reliable proxy for human-origin LLM-mediated writing. Performance is evaluated primarily at a conservative low-false-positive operating point, with global ranking separability as a secondary view, since false positives on human-authored or human-origin text can be costly in educational, scientific, and organizational settings~\citep{LIANG2023100779,website-new-ai-classifier-for-indicating-ai-written-text}.

The main contributions of this work are as follows:
\begin{itemize}[leftmargin=*]
    \item \textbf{Conceptual contribution:} a benchmark-transfer framing for AI-text detector evaluation, separating content origin from linguistic surface instead of pooling all LLM-involved text into one class.
    \item \textbf{Design contribution:} a matched four-regime authorship--surface design with \human{}, \freellm{}, \htol{}, and same-generator \llmtol{} across XSum, WritingPrompts, and OpenWebText.
    \item \textbf{Evaluation contribution:} block-level paired deltas, macro-averaging, bootstrap confidence intervals, and \tprfpr{} as the primary operating endpoint, alongside AUROC.
    \item \textbf{Empirical contribution:} evidence that strong detectors retain high low-FPR recall on \llmtol{} but degrade sharply on \htol{}, showing that the standard \human{} vs.\ \freellm{} benchmark overestimates robustness for human-origin LLM-mediated writing.
    \item \textbf{Diagnostic contribution:} textual transformation analysis and detector-, dataset-, and generator-level heterogeneity results showing that the operational \htol{}--\llmtol{} gap is associated with both source origin and transformation strength, and is not uniform across detector families.
\end{itemize}

\arb{} is designed to advance current AI-text detection benchmarking by jointly combining matched source items across four regimes, a same-generator \llmtol{} control, paired block-level deltas, a shared \human{} reference, and a benchmark-transfer framing evaluated under a strict low-FPR endpoint; Section~\ref{subsec:related_gap} positions this design against the closest prior benchmarks once they have been reviewed. Unlike paraphrase-robustness and humanization studies that report detector degradation under a single rewriting attack, \arb{} treats rewriting as two distinct, matched authorship--surface conditions and asks whether a benchmark built on one condition (\freellm{}) predicts performance on the other (\htol{}, \llmtol{}). This contributes toward a more precise account of when, and for whom, AI-text detectors remain reliable.

Section~\ref{sec:results} addresses these objectives directly: it first establishes baseline detectability under \human{} vs.\ \freellm{}, then quantifies degradation under \htol{} and under same-generator \llmtol{}, then compares the two LLM-mediated regimes to isolate the operational \htol{}--\llmtol{} gap, and finally characterizes heterogeneity across detector families, dataset domains, and generator models. Section~\ref{sec:discussion} interprets the resulting evidence in light of the objectives stated above.

\section{Related Work}
\label{sec:related_work}
This section positions the study within prior work on AI-text detection, rewriting robustness, benchmark design, and low-false-positive evaluation. The goal is not to survey all detector variants exhaustively, but to identify the empirical limitations of standard human-versus-LLM benchmarks and motivate the need for a matched four-regime evaluation.

\subsection{AI-text detection families}
\label{subsec:related_detection}

AI-text detection has been studied through several detector families. A recent survey organizes this space along a passive/active axis: passive detectors infer authorship post hoc from a text alone, whereas active approaches, principally watermarking and generation-log retrieval, require cooperation from the generation pipeline itself~\citep{XIANG2026}. \arb{} is restricted to passive, post-hoc detectors; watermarking is out of scope, as it targets a different deployment setting in which the detector controls or has privileged access to the generator. Within passive detection, statistical detectors such as GLTR expose token-rank irregularities~\citep{gehrmann2019gltr}; supervised detectors fine-tune encoders such as BERT or RoBERTa~\citep{ippolito2020automatic,liu2021roberta,li2024mage}; likelihood- and curvature-based zero-shot detectors use probability structure from a reference model, as in DetectGPT and FastDetectGPT~\citep{mitchell2023detectgpt,su2023fastdetectgpt}; contrastive zero-shot detectors such as Binoculars compare paired observer/performer likelihoods~\citep{hans2024binoculars}; and robustness-oriented supervised detectors such as RADAR train against adversarial paraphrasing~\citep{hu2023radar}. A newer rewriting-as-probe family instead uses an LLM's own rewrite or correction of a candidate text as the detection signal, exploiting the finding that LLMs edit already-LLM-generated text less than human-written text: RAIDAR measures the edit distance induced by an LLM rewrite of the input~\citep{mao2024raidar}; MAGRET instead uses rewrite similarity to detect and attribute authorship without log-probability access~\citep{huang-etal-2025-magret}; L2R fine-tunes the rewriter to amplify the RAIDAR edit-distance gap, improving cross-domain AUROC~\citep{hao-etal-2025-learning}; and GECScore scores similarity to a grammar-corrected version of the text, reporting robustness to cross-domain and paraphrase attacks~\citep{wu-etal-2025-wrote}. Recent work also explores more specialized zero-shot, interpretable, or domain-adaptive detectors, including inverse-prompt and distribution-alignment approaches~\citep{chen2025ipad,chen2025divscore}.

These detector families are not interchangeable: supervised encoders may learn dataset- or generator-specific cues and degrade under distribution shift; likelihood-based detectors depend on the fit between the reference model and the evaluated distribution; contrastive methods can be strong on direct generation but behave differently under rewriting; and robustness-oriented supervised detectors still need evaluation across domains, generators, and operating points. Two recent large-scale comparisons confirm this heterogeneity directly: one spanning classical, neural, fusion, and prompting-based detectors under domain and generator shift, with every family losing 5--30 AUROC points~\citep{baidya2026detectingmachinecomprehensivebenchmark}; the other fusing stylometric features with transformer representations, which improves cross-domain transfer but leaves backbone-specific failure modes~\citep{mady2026featureaugmentedtransformersrobustaitext}. For this reason, a benchmark should treat detectors as the objects of evaluation and report family-level patterns rather than only a single aggregate score.

\subsection{Paraphrase and humanization robustness}
\label{subsec:related_paraphrase}

Paraphrasing and humanization are established challenges for AI-text detection. \citet{krishna2023paraphrasing} showed that paraphrasing LLM-generated text can evade detectors while preserving semantics, and proposed retrieval as a defense; \citet{sadasivan2023reliably} argued that reliable detection is difficult under realistic attacks and transformations. Red-teaming studies show that LLM-assisted word substitution, style-changing prompts, or learned paraphrase policies can compromise detector performance~\citep{shi-etal-2024-red,Weichert2024DUPEDU,ranganath2026stealthrl}, and security-oriented evaluations of deepfake text detection find that defenses degrade under adaptive settings~\citep{10179387}. Three recent systems push this threat model further: TempParaphraser simulates high-temperature sampling through repeated normal-temperature rewrites~\citep{huang-etal-2025-tempparaphraser}; GradEscape trains a lightweight paraphraser against detector gradients or a query-extracted surrogate, evading deployed commercial detectors~\citep{meng2025gradescape}; and HUMPA applies a decoding-time logit shift from a preference-tuned proxy model that transfers across writing disciplines and languages~\citep{wang2025humanizing}. \citet{weberwulff2023testing} evaluate detection tools under machine translation and content obfuscation, showing that text transformations strongly affect tool reliability, and DAMAGE reports that many detectors struggle when processed by humanizer or paraphrasing systems~\citep{masrour2025damage}. Even without an adversarial framing, meaning-preserving transformation of AI-generated text alone weakens detection: on the semantic-invariant split of HC3 PLUS, where ChatGPT answers are translated, summarized, or paraphrased while their content is held fixed, fine-tuned encoders that reach near-ceiling balanced accuracy on untransformed text drop by roughly 12--13 points~\citep{mady2026featureaugmentedtransformersrobustaitext}.

These studies establish that rewriting and humanization can break detector-visible signals. They primarily frame the problem as detector degradation under adversarial or transformed input. Unlike attack-centered paraphrase benchmarks, \htol{} is not treated only as adversarial evasion. It is modeled as a plausible assisted-writing workflow in which human-origin content receives an LLM-mediated surface through polishing, rewriting, or assisted composition.

\subsection{Mixed authorship and paraphrase-origin benchmarks}
\label{subsec:related_mixed}

A complementary line of work constructs benchmarks that go beyond binary direct-generation detection: M4 evaluates multi-generator, multi-domain, multilingual settings~\citep{wang2023m4}; RAID targets robust evaluation under diverse attacks~\citep{dugan2024raid}; MAGE studies detection in the wild~\citep{li2024mage}; M4GT-Bench adds mixed human--machine detection and boundary localization~\citep{wang2024m4gtbench}; and DetectRL introduces real-world stressors including human revisions, writing errors, and mixing~\citep{wu2024detectrl}. Model rankings and apparent detector quality vary substantially across datasets, metrics, and protocols~\citep{prohl2024benchmarking,stowe2026spotlights}, a pattern echoed by two further large-scale benchmarks: a multi-family, cross-domain, cross-generator evaluation whose humanization protocol, like the paraphrase-attack studies above, rewrites only already-LLM-generated text rather than a human-origin condition comparable to \htol{}~\citep{baidya2026detectingmachinecomprehensivebenchmark}; and CUDRT, which pursues operational rather than authorship-origin diversity across a bilingual Create/Update/Delete/Rewrite/Translate taxonomy, finer-grained than \arb{}'s four regimes but not anchored to matched source items sharing a common \human{} baseline~\citep{10.1145/3779427}.

Two recent benchmarks are particularly close to the present study. \citet{lau2025effects} introduce the Human \& LLM Paraphrase Collection (HLPC), which explicitly combines human-written texts, LLM-generated texts, and their paraphrases, and reports performance at the 1\% FPR operating point for both types. PADBen distinguishes paraphrasing of human-authored content (authorship obfuscation) from paraphrasing of LLM-generated content (plagiarism evasion) and reports a performance asymmetry between the two cases~\citep{zha2025padben}. These works are the closest prior studies to \arb{}, and Table~\ref{tab:related_benchmarks} positions them explicitly against the present design.

HLPC and PADBen ask whether paraphrased or source-aware transformed texts remain detectable. \arb{} asks a different evaluation-validity question: whether detector performance estimated on direct LLM generation transfers to matched LLM-mediated regimes under a shared human reference distribution. The distinction is in the benchmark structure and estimand, rather than in the presence of paraphrased text alone.

\begin{table*}[t]
\centering
\small
\renewcommand{\arraystretch}{1.35}
\caption{Structural comparison of the closest benchmark designs. ``Partial'' indicates that a feature is present only for part of a benchmark or not used as the central evaluation design.}
\label{tab:related_benchmarks}
\begin{tabularx}{\textwidth}{Xcccc}
\toprule
Feature & M4/RAID/MAGE & HLPC & PADBen & \arb{} \\
\midrule
Matched source item & No & Partial & Partial & Yes \\
Four-regime quartet & No & No & No & Yes \\
\human{} baseline reused across tasks & No & Partial & Partial & Yes \\
Same-generator \llmtol{} control & No & No & No & Yes \\
Paired target-minus-baseline deltas & No & No & No & Yes \\
Low-FPR endpoint as primary metric & Partial & Yes & Partial & Yes \\
Benchmark-transfer framing & No & No & No & Yes \\
Transformation diagnostics & Partial & Partial & Partial & Yes \\
\bottomrule
\end{tabularx}
\end{table*}

A related line of work audits the assumptions encoded in mixed-authorship datasets themselves, rather than detector robustness to a fixed set of transformations. \citet{dycke2026yourai} formalize AI-text detection tasks, or ``notions,'' along three axes---the normative standard for acceptable AI use, the granularity of the human--AI genesis (document-, boundary-, or sentence-level), and the attacker model---and show, via AITDNA, a dataset of naturally logged human--LLM co-writing sessions, that synthetic benchmarks such as DetectRL, Mixset, SenDetEx, and BD misrepresent natural co-creation in AI-token proportion, boundary count, and human--AI linguistic gap. Their document-level notion is governed by an explicit parameter $\tau$, the minimum AI-token share for a document-level AI label; leaving $\tau$ implicit, as direct-generation datasets effectively do by labeling any AI involvement as positive, produces incompatible evaluation targets. This formalizes, at the dataset-construction level, the same concern that motivates \arb{}'s regime design: collapsing content origin and linguistic surface into one label obscures what a detector is asked to recognize. The two studies are complementary in what they hold fixed: \citeauthor{dycke2026yourai} vary the notion applied to a fixed corpus of natural writing to expose hidden dataset assumptions, while \arb{} fixes a single notion (content origin vs.\ LLM-mediated surface) and varies the rewriting regime under matched source items to test benchmark transfer; their analysis does not include a same-generator second-pass control or paired block-level deltas.

Overall, prior benchmarks cover important dimensions such as domain diversity, mixed authorship, paraphrase robustness, humanization, and---in the case of AITDNA---the realism of human--AI co-writing traces underlying dataset construction. To the best of our knowledge, existing benchmarks do not jointly combine matched source items across four regimes, a same-generator \llmtol{} control, paired block-level deltas, a shared \human{} reference, and a benchmark-transfer framing evaluated under a strict low-FPR endpoint.
Prior work also shows that aggregate ranking metrics are not sufficient
for evaluating AI-text detectors in settings where false positives are
costly. Several studies therefore report detector performance at fixed
low false-positive rates, including 1\% FPR or stricter operating
points~\citep{krishna2023paraphrasing,lau2025effects,masrour2025damage,ayoobi2025shield,chen2025divscore,ranganath2026stealthrl}. 
This motivates our use of \tprfpr{} alongside AUROC in the evaluation protocol.
\subsection{Remaining gap: matched benchmark transfer}
\label{subsec:related_gap}

The comparison in Table~\ref{tab:related_benchmarks} shows that existing benchmarks address several adjacent problems: detector robustness, mixed-authorship detection, paraphrase effects, and humanization attacks.
The remaining gap is narrower but important: whether performance measured on the conventional \human{} vs.\ \freellm{} benchmark transfers to matched LLM-mediated rewriting regimes under a shared human reference distribution.

\arb{} addresses this gap by anchoring all regimes to matched source items and by using the same-generator \llmtol{} as a controlled second-pass condition rather than another paraphrase attack. This design enables paired comparisons among direct LLM generation, human-origin LLM rewriting, and LLM-origin second-pass rewriting within the same dataset--generator blocks, rather than treating rewritten samples as unrelated pooled positives, and thereby tests whether detectors that perform well under conventional \human{} vs.\ \freellm{} evaluation retain low-FPR recall under matched \htol{} and \llmtol{} regimes.

\section{\arb{} Benchmark Design}
\label{sec:benchmark_design}
\begin{figure}[pos=ht!]
\centering
\resizebox{0.8\textwidth}{!}{\begin{tikzpicture}[
    >=Latex,
    font=\footnotesize,
    node distance=5mm and 9mm,
    datastore/.style={
    draw,
    thick,
    rectangle,
    fill=baseColor,
    align=center,
    text width=3.1cm,
    minimum height=8mm,
    double,
    double distance=1pt
    },
    source/.style={datastore},
process/.style={draw, rounded corners, fill=white, align=center, text width=3.2cm, minimum height=7mm},
    regime/.style={draw, rounded corners, thick, fill=gray!7, align=center, text width=2.7cm, minimum height=7mm},
    result/.style={draw, rounded corners, thick, fill=white, align=center, text width=4.1cm, minimum height=7mm},
    arrow/.style={->, semithick}
]

\node[source] (datasets) {Source datasets};
\node[process, below=of datasets] (filtering) {Filtering and\\length-stratified sampling};
\node[source, below=of filtering] (matched) {Matched source item};

\node[regime, below left=7mm and 37mm of matched] (human) {\human{}};
\node[process, below=7mm of matched] (topic) {Topic extraction};
\node[process, below right=7mm and 37mm of matched] (h2lprompt) {\htol{} rewrite prompt};

\node[process, below=of topic] (freeprompt) {\freellm{} generation prompt};
\node[regime, below=of h2lprompt] (h2l) {\htol{}};
\node[regime, below=of freeprompt] (freellm) {\freellm{}};
\node[process, right=13mm of freellm] (llm2lprompt) {\llmtol{} rewrite prompt\\same generator};
\node[regime, below=of llm2lprompt] (llm2l) {\llmtol{}};

\node[result, below=60mm of matched] (valid) {Valid matched block};
\node[process, below=of valid] (scoring) {Detector scoring};

\draw[arrow] (datasets) -- (filtering);
\draw[arrow] (filtering) -- (matched);
\draw[arrow] (matched) -- (human);
\draw[arrow] (matched) -- (topic);
\draw[arrow] (matched) -- (h2lprompt);
\draw[arrow] (topic) -- (freeprompt);
\draw[arrow] (freeprompt) -- (freellm);
\draw[arrow] (h2lprompt) -- (h2l);
\draw[arrow] (freellm) -- (llm2lprompt);
\draw[arrow] (llm2lprompt) -- (llm2l);

\draw[arrow] (human.south) |- (valid.west);
\draw[arrow] (freellm.south) -- (valid.north);
\draw[arrow] (h2l.south) |- (valid.east);
\draw[arrow] (llm2l.south) |- (valid.east);
\draw[arrow] (valid) -- (scoring);

\end{tikzpicture}
}
\caption{End-to-end \arb{} construction and evaluation workflow. A filtered, length-stratified human source item anchors four matched regimes. \human{} retains the source text; \freellm{} is generated from a dataset-specific topic; \htol{} rewrites the human source; and \llmtol{} rewrites the corresponding \freellm{} output with the same generator. Only complete validated quartets proceed to detector scoring and paired block-level analysis.}
\label{fig:arb-construction-pipeline}
\end{figure}
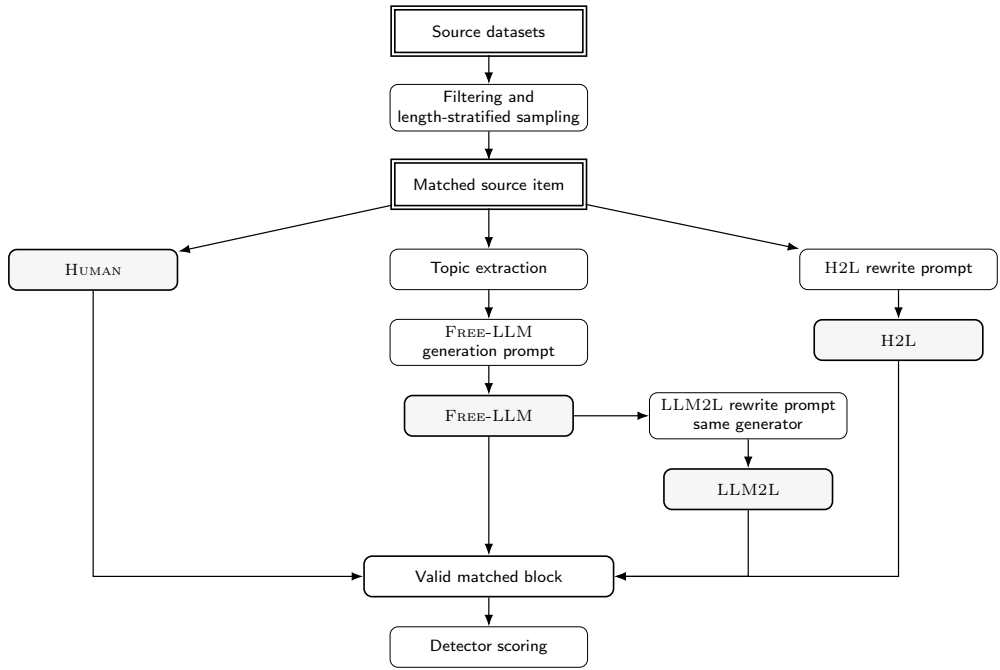

This section describes the end-to-end construction of \arb{}. Figure~\ref{fig:arb-construction-pipeline} summarizes the transformation flow used to generate the benchmark dataset: each sampled human source item anchors the \human{} condition, provides the topic basis for \freellm{} generation, is rewritten to produce \htol{}, and links the corresponding \freellm{} output to the same-generator \llmtol{} rewrite. The pipeline starts from the training splits of three Hugging Face datasets (XSum, WritingPrompts, and OpenWebText) applies source preprocessing, and draws a seeded stratified random sample of 600 human texts from each dataset. Each sampled text is assigned a stable matched-source identity and then follows three paths: the original text is retained as \human{}; a dataset-specific topic is passed to a generation prompt to create \freellm{}; and the original human text is passed to a rewrite prompt to create \htol{}. The resulting \freellm{} text is subsequently rewritten by the same generator to create \llmtol{}. Detectors are then applied to every regime variant, after which detection metrics are computed. 

\subsection{Four-regime design}
\label{subsec:four_regime_design}

The benchmark operationally contrasts content-origin and LLM-mediated surface regimes under a matched benchmark-transfer design. Table~\ref{tab:regimes} defines the four regimes. \human{} is the reference human condition. \freellm{} is the baseline machine-generation condition. \htol{} and \llmtol{} are experimental conditions in which an LLM mediates the final surface through rewriting.

\begin{table}[t]
\centering
\caption{Four-regime benchmark design.}
\label{tab:regimes}
\begin{tabular*}{\tblwidth}{@{} LLL@{}}
\toprule
Regime & Content origin & Linguistic surface \\
\midrule
\human{} & human & human-written \\
\freellm{} & LLM & LLM-generated \\
\htol{} & human & LLM-mediated rewrite \\
\llmtol{} & LLM & LLM-mediated rewrite \\
\bottomrule
\end{tabular*}
\end{table}

For each source sample, \human{} is the original human text. \freellm{} is generated from a topic derived from the same source item. \htol{} is produced by rewriting the human source text. \llmtol{} is produced by applying the same rewriting protocol to the corresponding \freellm{} text. \freellm{} is not intended to be a semantic paraphrase of the \human{} source; it represents the conventional topic-conditioned direct-generation baseline. \htol{} and \llmtol{} instead instantiate rewriting regimes. Comparisons are therefore interpreted as benchmark-transfer contrasts rather than semantic-equivalence contrasts across all four regimes. In the main experiment, \llmtol{} is generated by the same model that produced the corresponding \freellm{} text. This same-generator design avoids adding a crossed generator--rewriter factor and makes \llmtol{} a controlled second-pass condition. It tests whether a second LLM-mediated surface pass, by itself, makes LLM-origin text resemble the harder \htol{} condition.

\subsection{Datasets and domains}
\label{subsec:datasets}

We used three English datasets selected to represent distinct textual domains: XSum for news and factual writing~\citep{narayan-etal-2018-dont}, WritingPrompts for creative and narrative writing~\citep{fan-etal-2018-hierarchical}, and OpenWebText for web/general writing~\citep{Gokaslan2019OpenWeb}.

XSum is a dataset for evaluating summarization models. Each dataset entry has a ``document'' component that provides detailed narrative information and a ``summary'' component that captures the key points.
WritingPrompts is a large FAIR dataset containing human-written stories paired with prompts from an online forum.
OpenWebText is an open-source replication of the WebText dataset from OpenAI provided by Aaron Gokaslan. The dataset has been created by extracting URL links from the ``Reddit submission dataset'' and parsing the HTML pages of related Reddit posts.

The topic used in the prompt template to generate the \freellm{} texts was dataset-dependent (Table~\ref{tab:datasets}): the summary field for XSum, the prompt for WritingPrompts, and the first two cleaned sentences from the text column, with a maximum of 40 words for OpenWebText.  For exact replication, the sources were loaded with the Hugging Face \texttt{datasets} library using the \texttt{train} split in all three cases. Dataset-specific identifiers and fields are reported in Table~\ref{tab:datasets}. The use of multiple domains is a validity control against conclusions driven by a single genre or source distribution.

\begin{table}[t]
\centering
\caption{Dataset-specific source fields and topic extraction for \freellm{} generation.}
\label{tab:datasets}
\begin{tabularx}{\linewidth}{lXll}
\toprule
Dataset & Hugging Face identifier & Human-text field & Topic for \freellm{} \\
\midrule
XSum & \texttt{EdinburghNLP/xsum} & \texttt{document} & \texttt{summary} \\
WritingPrompts & \texttt{euclaise/}\allowbreak\texttt{writingprompts} & \texttt{story} & \texttt{prompt} \\
OpenWebText & \texttt{Skylion007/}\allowbreak\texttt{openwebtext} & \texttt{text} & first two cleaned sentences, at most 40 words \\
\bottomrule
\end{tabularx}
\end{table}

\subsection{Source preprocessing and stratified random sampling}
\label{subsec:length_stratification}
Detector performance can depend on the amount of available text. We therefore restricted the eligible source pool to texts of 150--500 whitespace-delimited words and removed source items rejected by the implemented structural-artifact and content-suitability preprocessing. Each eligible human text was assigned to one of three length strata: short texts contained 150--220 words, medium texts 221--350 words, and long texts 351--500 words. For each dataset, we then used stratified random sampling with seed 42 to draw 200 rows from each stratum. The resulting sample contains exactly 600 rows per dataset, i.e., 1,800 human source rows in total.

\subsection{Generator models}
\label{subsec:generator_models}

The benchmark uses four open-weight instruction-tuned generator families. We focus on open-weight generators to ensure reproducibility and release-compatible benchmarking. The resulting estimates should not be generalized to closed-source proprietary systems without additional evaluation. For each source-item--generator pair, the selected model creates the \freellm{} text, rewrites the human source as \htol{}, and performs the same-generator second pass from \freellm{} to \llmtol{}. 

All four models were loaded in \texttt{bfloat16} precision and received the same fixed system prompt and the same fixed regime-specific user templates. No model-specific prompt wording or prompt tuning was used. For each source item, only the prompt template variables were instantiated (Section~\ref{subsec:prompt_protocol}): the topic and stratum bounds for \freellm{}, the human source text for \htol{}, and the corresponding \freellm{} output for \llmtol{}.  Table~\ref{tab:generator_models} therefore reports only model-specific information. Model-family references are provided where a stable technical report or model paper is available~\citep{grattafiori2024llama3herdmodels,gemmateam2024gemmaopenmodelsbased,mistral7b,qwen25technicalreport}.

\begin{table}[t]
\centering
\caption{Generator models and exact Hugging Face identifiers.}
\label{tab:generator_models}
\begin{tabularx}{\linewidth}{lXr}
\toprule
Alias & Hugging Face model identifier & Parameters \\
\midrule
llama32\_3b & \texttt{meta-llama/Llama-3.2-3B-Instruct} & 3B \\
qwen25\_7b & \texttt{Qwen/Qwen2.5-7B-Instruct} & 7B \\
mistral7b & \texttt{mistralai/Mistral-7B-Instruct-v0.3} & 7B \\
gemma2\_9b & \texttt{google/gemma-2-9b-it} & 9B \\
\bottomrule
\end{tabularx}
\end{table}

The decoding configuration was specified before detector evaluation and was not tuned post hoc to maximize detector degradation. The design objective was methodological rather than adversarial: to construct benchmark texts that were sufficiently diverse to avoid trivially templatic outputs, while remaining coherent, semantically faithful, and comparable across generator families. This choice follows prior work showing that sampling decisions mediate a quality--diversity trade-off in generated text and can materially affect downstream evaluation conclusions~\citep{ippolito2020automatic,chung-etal-2023-increasing,zhou2024balancing}. Accordingly, we adopted a moderate stochastic decoding regime rather than deterministic decoding or aggressively high-temperature sampling, together with the fixed task and output constraints described above.

To ensure transparent and reproducible generation, we fix the decoding settings across models and regimes. Generation uses the Hugging Face Transformers backend with \texttt{do\_sample=True}, \texttt{temperature}=0.7, \texttt{top\_p}=0.9, \texttt{top\_k}=40, and \texttt{max\_new\_tokens}=512. The global random seed is fixed to 42 for sampling and bootstrap evaluation. Prompts are rendered through the model chat template before decoding. For Gemma-family models, which did not use the same system-role template in our implementation, the system instruction is merged into the user prompt to preserve a consistent output-only generation policy.

\subsection{Prompt design}
\label{subsec:prompt_protocol}
This section explains the rationale behind the prompt design, while Appendix~\ref{app:prompts} reports the prompt templates verbatim.

After sampling, the human text, its source-dataset identifier, its original source index, and its length band defined the matched source item. 
Each source item was used to generate texts using the generator models described in Section~\ref{subsec:generator_models}.

Prompt wording can materially affect model behavior. We therefore treated it as a controlled component of the benchmark rather than tuning it separately by dataset or model~\citep{zhou2023humanlevel,sahoo2024systematic,schulhoff2024promptreport}. 

Prompts are composed of two parts: a fixed system prompt and a parametrized task-dependent prompt.

They are designed to be short, explicit, neutral in style, and aligned with the three operations being studied. 
We deliberately avoided instructions such as ``humanize,'' named stylistic personas, or detector-evasion objectives, because these would introduce a separate adversarial or stylistic manipulation.

The minimal system prompt aims to reduce prompt-induced formatting artifacts. It asks every model to return only the requested text and suppresses prefaces, explanations, headings, and Markdown. This reduces output-format artifacts that are unrelated to the authorship regime.
\paragraph{\freellm{} template.}
For each matched source item and generator model, the \freellm{} template requests fluent, self-contained English text on the extracted topic, requires original wording and structure, and prohibits references to the source text, headings, and bullet points.

Topic extraction is dataset-specific and deterministic: XSum uses the dataset's \texttt{summary} field, WritingPrompts uses its \texttt{prompt} field, and OpenWebText uses the first two cleaned sentences, falling back to the first sentence and truncating it to 40 words when necessary. The topic is an input only to the \freellm{} branch, while the complete human text is the input to the \htol{} branch. \texttt{min\_words} and \texttt{max\_words} variables are also set to the boundaries of the sampled source's length stratum (150--220, 221--350, or 351--500 words). Thus, length is controlled from the pre-generation stratum without supplying the human article itself as generation content.

\paragraph{\htol{} template.}
Independently, the \htol{} template supplies the complete human source and requests a fluent rewrite that (i) preserves meaning, factual claims, entities, and relationships, (ii) adds no new information, (iii) removes no important information, (iv) changes wording and sentence structure where possible, (v) and remains approximately the same length.

\paragraph{\llmtol{} template.}

The \llmtol{} template applies an analogous meaning-preserving rewrite instruction to the corresponding \freellm{} output. The same model that created \freellm{} performs this second pass, so \human{} $\rightarrow$ \htol{} and \freellm{} $\rightarrow$ \llmtol{} differ in input-content origin while retaining a closely parallel rewriting operation. Across all datasets and generator families, the templates remain fixed: only the topic, source text, generated text, and the stratum-derived length variables are instantiated.

\subsection{Matched blocks and comparison readiness}
\label{subsec:valid_blocks}

A matched unit is defined at the level of \texttt{sample\_id $\times$ generator\_model}. It contains the unchanged \human{} text and the \freellm{}, \htol{}, and \llmtol{} outputs associated with the same one of the 1,800 sampled source items and generator model. Source rows are not replaced or resampled after generation; the fixed stratified sample is the basis for all three prompt applications. Balance by dataset and human-source length stratum therefore originates in the initial sampling step rather than in post-generation selection.

Each generation call was allotted up to three retry attempts, but no call exhausted this budget: every one of the 7,200 attempted source-item--generator combinations (1,800 source items $\times$ 4 generators) produced a complete quartet, so all 7,200 quartets (23,400 individual texts; Table~\ref{tab:dataset_stats}) entered detector scoring and paired analysis. Beyond the prompt-level constraints described in Section~\ref{subsec:prompt_protocol} (the stratum-derived target length for \freellm{}, and the meaning-preservation and approximate-length instructions for \htol{} and \llmtol{}), no post-hoc semantic, length, or language filtering was applied to the released benchmark. Retention was therefore not conditioned on output quality, and a small fraction of generated texts deviate from the intended target profile: automated checks on the released dataset identify 0.64\% of generated texts shorter than 50 words, 0.08\% matching a refusal or policy-disclaimer pattern (e.g., ``I can't fulfill this request.''), 0.03\% flagged as non-English by automatic language identification, and 0.01\% exact duplicates; these categories overlap, and their union covers 0.67\% (145/21,600) of generated texts. These cases are retained in the released dataset rather than silently dropped, and can be identified and excluded using the released text and word-count metadata. Section~\ref{sec:limitations} discusses the resulting limitation.

The retained identifiers provide the joins required for replication: \texttt{sample\_id} links the four regime variants, the source index traces the item to the sampled dataset, and the generator identifier records the model used for all generated variants in that block. The resolved experiment configuration fixes dataset names and splits, preprocessing settings, length bands, random seed, prompt templates, decoding parameters, and output paths. These artifacts separate benchmark construction from detector evaluation and permit the complete quartet to be reconstructed before any detector score is inspected.

Matched comparisons are subsequently computed within the same dataset $\times$ generator block. For example, $\Delta$\tprfpr{} for \htol{} compares \human{} vs. \htol{} with \human{} vs. \freellm{} under matched block conditions, before the paired deltas are macro-averaged across blocks. The same alignment is used for \llmtol{} and for the direct \llmtol{}--\htol{} source-origin comparison. This avoids comparisons between unrelated pooled samples and carries the construction-stage matching into statistical estimation. The matched blocks then enter the detector-scoring procedure defined in Section~\ref{sec:evaluation_protocol}.

\subsection{Dataset description and final statistics}
\label{subsec:dataset_description}

\arb{} is an English-language text collection of 23,400 samples: 1,800 \human{} source texts (600 per dataset; Section~\ref{subsec:length_stratification}) and, for each of the four generator models in Table~\ref{tab:generator_models}, one \freellm{}, \htol{}, and \llmtol{} text per source item, i.e., 7,200 texts per generated regime (1,800 source items $\times$ 4 generators). As reported in Section~\ref{subsec:valid_blocks}, every attempted source-item--generator combination is complete, so these figures also describe the exact set of matched quartets used in the paired analyses of Sections~\ref{sec:results} and \ref{sec:discussion}.

Table~\ref{tab:dataset_stats} reports the final composition by source dataset and regime, together with word-count statistics computed on whitespace-delimited tokens. \human{} texts are length-stratified by construction and therefore fall within 150--500 words in every dataset. \freellm{}, \htol{}, and \llmtol{} texts are only softly constrained by the generation and rewriting prompts (a target stratum for \freellm{}, an approximate-length instruction for the rewrites; Section~\ref{subsec:prompt_protocol}) and consequently show a wider spread around a lower mean, with \llmtol{} texts on average the shortest as a result of two successive LLM-mediated passes.

\begin{table}[t]
\centering
\caption{Final composition of \arb{} by source dataset and regime. Each cell reports the number of texts and, in parentheses, the mean $\pm$ standard deviation word count. Within every dataset $\times$ regime cell, texts are balanced exactly across the four generator models (Table~\ref{tab:generator_models}); \human{} texts are shared across generator blocks and are therefore not multiplied by generator.}
\label{tab:dataset_stats}
\begin{tabular*}{\tblwidth}{@{} LCCCC@{}}
\toprule
Dataset & \human{} & \freellm{} & \htol{} & \llmtol{} \\
\midrule
XSum & 600 (295$\pm$102) & 2400 (234$\pm$105) & 2400 (232$\pm$72) & 2400 (204$\pm$96) \\
WritingPrompts & 600 (298$\pm$104) & 2400 (254$\pm$107) & 2400 (240$\pm$76) & 2400 (229$\pm$99) \\
OpenWebText & 600 (297$\pm$103) & 2400 (224$\pm$107) & 2400 (209$\pm$72) & 2400 (196$\pm$97) \\
\midrule
Total & 1800 (297$\pm$103) & 7200 (237$\pm$107) & 7200 (227$\pm$75) & 7200 (210$\pm$99) \\
\bottomrule
\end{tabular*}
\end{table}

\textbf{Licensing.} XSum source articles are BBC news text; the canonical Hugging Face release (\texttt{EdinburghNLP/xsum}) does not declare an explicit dataset license and is distributed by its authors for research use~\citep{narayan-etal-2018-dont}. WritingPrompts originates from user-submitted posts on Reddit's r/WritingPrompts, compiled by \citet{fan-etal-2018-hierarchical}; the Hugging Face mirror used for sourcing (\texttt{euclaise/writingprompts}) is tagged with the MIT license. OpenWebText packaging is released under a CC0 dedication by its curators, who do not claim ownership of the underlying scraped web text and provide a notice-and-takedown mechanism for copyright holders~\citep{Gokaslan2019OpenWeb}. \arb{} itself is released under the Apache License 2.0 (Section~\ref{sec:data_availability}); as stated there, downstream users remain responsible for the licensing terms of the three source corpora summarized above.

\section{Evaluation Protocol}
\label{sec:evaluation_protocol}
This section defines how detectors are evaluated on the benchmark. We describe the detector families, primary metrics, aggregation and bootstrapping sampling strategies, and textual transformation diagnostics.

\subsection{Detectors}
\label{sec:detectors}

The objects of evaluation are AI-text detectors representing complementary detector families. We include supervised encoder baselines, a paraphrase-robust supervised detector, a zero-shot likelihood/curvature detector, and a zero-shot contrastive likelihood detector. \arb{} is not intended to rank detectors exhaustively. Its purpose is to test whether detector families preserve performance ordering and low-FPR recall when moving from direct-generation evaluation to matched rewriting regimes.

\begin{table}[t]
\centering
\small
\caption{Detector families included in the benchmark and rationale for inclusion.}
\label{tab:detectors}
\begin{tabularx}{\linewidth}{@{}p{0.17\linewidth}p{0.16\linewidth}p{0.16\linewidth}X@{}}
\toprule
Detector & Family & Setting & Role \\
\midrule
BERT-Defense & Supervised encoder & Trained & Encoder baseline~\citep{10179387} \\
RoBERTa-Defense & Supervised encoder & Trained & RoBERTa encoder baseline~\citep{10179387,liu2021roberta} \\
RADAR & Robust supervised & Trained & Paraphrase-robust detector~\citep{hu2023radar} \\
Binoculars-falcon-7b & Contrastive likelihood & Zero-shot & Likelihood-ratio detector~\citep{hans2024binoculars} \\
FastDetectGPT & Likelihood/curvature & Zero-shot & DetectGPT-style detector~\citep{mitchell2023detectgpt,su2023fastdetectgpt} \\
\bottomrule
\end{tabularx}
\end{table}

Detector configurations followed the released implementations or paper-recommended settings whenever available. Hardware-driven adjustments were applied only when needed to run the experiments on the available workstation, for example, through precision, quantization, or maximum observed sequence length. These choices were made for execution feasibility and were not tuned on \arb{} labels or on regime-specific performance. Each detector configuration was kept fixed across all datasets, generator models, and regimes, avoiding detector retuning as a confounding factor in the comparison among \freellm{}, \htol{}, and \llmtol{}.

For BERT-Defense and RoBERTa-Defense, we used the pretrained checkpoints from the experimental setup of \citet{10179387}. These encoder baselines are not included as state-of-the-art competitors but as representative supervised detectors whose behavior under distribution shift provides a lower-bound comparison against zero-shot and robustness-oriented methods.

RADAR was included as a supervised detector explicitly designed for robustness to adversarial paraphrasing. We used the released checkpoint without additional training or task-specific adaptation. Since RADAR follows a detector-specific score convention in its released implementation, its output orientation was standardized before metric computation.

For Binoculars-falcon-7b, we used the Falcon-7B observer and Falcon-7B-Instruct performer configuration following the released Binoculars setup. Since the released Binoculars decision rule assigns AI-generated labels to scores below its threshold, its raw score has a lower-is-more-AI orientation. We retained the raw scores for reproducibility but used a sign-reversed score for AUROC and \tprfpr{} computation, so that larger standardized scores always correspond to stronger evidence for the positive class. No detector-specific threshold was tuned on \arb{}.

For FastDetectGPT, we used a fixed sampling/scoring model pair for all evaluated generators. We did not change the detector backbone according to the generator model, because doing so would introduce a confounding factor between generator identity and detector configuration. Following the updated recommendation of the official FastDetectGPT repository, we used Llama3-8B as the sampling model and Llama3-8B-Instruct as the scoring model. To make the paired zero-shot detector feasible on the available 32 GB GPU setup, inference was run with 8-bit quantization. This was an execution constraint, not a detector calibration step; no FastDetectGPT component was fine-tuned or threshold-tuned on \arb{}.

For every detector, raw scores and standardized scores were stored. AUROC and \tprfpr{} were computed from standardized continuous scores, not from detector-specific default labels. Detector-specific default thresholds were retained only for reproducibility and auxiliary inspection. This standardization ensures that all pairwise comparisons use a common score orientation, with \human{} as the negative class and the target regime as the positive class.

\subsection{Experimental setup}
\label{subsec:experimental_setup}

All experiments were conducted on a Linux workstation running Ubuntu 24.04.4 LTS with kernel 6.8.0-124-generic. The machine was equipped with an Intel Core Ultra 9 285K CPU, 62 GiB of RAM, and a 1.9 TB Samsung NVMe SSD formatted with ext4. GPU-based experiments were executed on a single NVIDIA GeForce RTX 5090 with 32 GB of VRAM, using NVIDIA driver 590.48.01.

The software environment used Python 3.12.3 and PyTorch 2.11.0+cu128. CUDA was available through PyTorch 12.8 with cuDNN 9.1.9, while the NVIDIA-SMI interface reported CUDA compatibility version 13.1. Detector configurations were therefore selected to be executable under this single-GPU setup while remaining fixed across regimes, datasets, and generator models.

\subsection{Primary metrics}
\label{subsec:primary_metrics}

We use \tprfpr{} as the primary operating metric. It is defined as the
fraction of positive examples detected when the false-positive rate on the
human class is constrained to 1\%. This conservative endpoint is important
because false accusations of AI authorship can create fairness and
accountability concerns, including biases against non-native English
writers~\citep{LIANG2023100779}.

We report AUROC as a complementary secondary metric. AUROC measures global
ranking separability: the probability that a randomly selected positive
example receives a higher machine-likeness score than a randomly selected
human example. It is threshold-independent and useful across the full score
range, but it does not indicate whether a detector retains useful recall
under the prespecified low-FPR constraint.

For each detector, dataset, and generator, we evaluated three matched
binary detection tasks: \human{} vs.\ \freellm{}, \human{} vs.\ \htol{},
and \human{} vs.\ \llmtol{}. In each task, \human{} texts are the negative
class and the target-regime texts are the positive class. The same \human{}
set is reused across the three tasks within each block to keep the human
reference distribution fixed. This prevents changes in the human negative
class from confounding comparisons among \freellm{}, \htol{}, and
\llmtol{}.

In the interpretation of results, we therefore lead with \tprfpr{} and use
AUROC as a secondary view of global separability. This distinction is
essential because detectors can retain non-trivial AUROC while having
near-zero low-FPR recall.

\subsection{Block-level estimation, aggregation, and uncertainty}
\label{subsec:block_level_estimation}

All performance estimates were first computed within each dataset $\times$ generator block. For each detector and block, we computed \tprfpr{} and AUROC for three binary comparisons: \human{} vs. \freellm{}, \human{} vs. \htol{}, and \human{} vs. \llmtol{}.

To quantify transfer from the conventional direct-generation benchmark, we computed paired target-minus-\freellm{} deltas within each block. A negative delta means that the detector performed worse in the target regime than in the \freellm{} baseline under the same dataset and generator. These deltas were computed separately for \htol{} and \llmtol{}. To compare the two LLM-mediated regimes directly, we computed an operational \htol{}--\llmtol{} gap as \llmtol{} minus \htol{} within the same block. A positive gap means that \llmtol{} was more detectable than \htol{} under the same dataset and generator.

Detector-level results were then obtained by macro-averaging across dataset $\times$ generator blocks. This gave each block equal weight and prevented larger or easier blocks from dominating the aggregate estimate. Deltas and operational gaps were computed within blocks before macro-averaging, rather than as differences between pooled averages.

We reported 95\% confidence intervals using a block-structured bootstrap with 5,000 resamples and random seed 42. Within each dataset $\times$ generator block, rows were resampled with replacement, and \tprfpr{} and AUROC were recomputed for each replicate. For deltas and operational gaps, the aligned rows across matched regimes were resampled jointly, preserving the pairing between regimes. Confidence intervals were reported as percentile intervals over the resulting bootstrap distribution. These intervals quantify uncertainty in the aggregate estimates, and heterogeneity across domains and generators is reported separately using heatmaps and stratified diagnostics.

\subsection{Textual transformation diagnostics}
\label{subsec:semantic_quality}

We computed textual transformation diagnostics for the two rewriting paths. For \htol{}, the source text was the original \human{} text and the target text was its LLM-mediated rewrite. For \llmtol{}, the source text was the corresponding \freellm{} output, and the target text was the same-generator second-pass rewrite.

For each source--target pair, we measured word ratio, token-level normalized edit distance (NED), lexical overlap using Jaccard similarity, and semantic similarity. These diagnostics were not used as detector inputs. They were used only to characterize how much each rewriting path changed its source text.

\section{Results}
\label{sec:results}
This section reports the benchmark results in order of the research objectives. We first establish baseline detectability under the conventional \human{} vs. \freellm{} condition, then quantify degradation under \htol{} and \llmtol{}, compare the two LLM-mediated regimes, and finally analyze heterogeneity across detectors, datasets, generators, and textual transformation features.

\subsection{Baseline detectability under \human{} vs. \freellm{}}
\label{subsec:rq1_baseline}

This first analysis evaluates the standard detector benchmark: distinguishing \human{} texts from directly generated \freellm{} texts. Table~\ref{tab:rq1_pairwise} reports \tprfpr{} and AUROC for all regimes; the \freellm{} rows are the baseline control.

At the primary operating point, FastDetectGPT and Binoculars-falcon-7b obtain the strongest baseline \tprfpr{} values, $0.912$ and $0.935$, respectively; RADAR reaches $0.668$. Their secondary AUROC values are also high ($0.990$, $0.983$, and $0.913$). In contrast, RoBERTa-Defense has near-zero \tprfpr{} ($0.019$) despite moderate AUROC ($0.569$), while BERT-Defense has negligible \tprfpr{} ($0.001$) and very low AUROC ($0.210$).

These results establish that direct LLM generation is detectable by the strongest zero-shot detectors and by RADAR, but not by all detector families at a strict low-FPR operating point. The baseline therefore provides a necessary control for interpreting subsequent robustness losses.

\begin{table*}[t]
\centering
\small
\caption{Pairwise detection performance across regimes, with the primary low-FPR endpoint reported first. Values are macro-averaged block-level estimates with 95\% confidence intervals across dataset $\times$ generator blocks. Bold values indicate the best score for each comparison and metric.}
\label{tab:rq1_pairwise}
\begin{tabularx}{\textwidth}{llXX}
\toprule
Detector & Comparison & \tprfpr{} & AUROC \\
\midrule
BERT-Defense & \human{} vs. \freellm{} & 0.001 [0.000, 0.003] & 0.210 [0.203, 0.217] \\
BERT-Defense & \human{} vs. \htol{} & 0.012 [0.008, 0.015] & 0.453 [0.444, 0.463] \\
BERT-Defense & \human{} vs. \llmtol{} & 0.002 [0.001, 0.004] & 0.233 [0.226, 0.240] \\
\midrule
RoBERTa-Defense & \human{} vs. \freellm{} & 0.019 [0.011, 0.030] & 0.569 [0.560, 0.578] \\
RoBERTa-Defense & \human{} vs. \htol{} & 0.025 [0.016, 0.038] & 0.586 [0.577, 0.595] \\
RoBERTa-Defense & \human{} vs. \llmtol{} & 0.016 [0.009, 0.025] & 0.538 [0.528, 0.547] \\
\midrule
FastDetectGPT & \human{} vs. \freellm{} & 0.912 [0.892, 0.929] & \textbf{0.990 [0.989, 0.992]} \\
FastDetectGPT & \human{} vs. \htol{} & \textbf{0.308 [0.264, 0.347]} & \textbf{0.874 [0.868, 0.879]} \\
FastDetectGPT & \human{} vs. \llmtol{} & 0.783 [0.754, 0.808] & \textbf{0.971 [0.968, 0.973]} \\
\midrule
Binoculars-falcon-7b & \human{} vs. \freellm{} & \textbf{0.935 [0.924, 0.942]} & 0.983 [0.980, 0.985] \\
Binoculars-falcon-7b & \human{} vs. \htol{} & 0.151 [0.135, 0.166] & 0.666 [0.658, 0.674] \\
Binoculars-falcon-7b & \human{} vs. \llmtol{} & \textbf{0.830 [0.816, 0.842]} & 0.952 [0.948, 0.955] \\
\midrule
RADAR & \human{} vs. \freellm{} & 0.668 [0.636, 0.694] & 0.913 [0.909, 0.918] \\
RADAR & \human{} vs. \htol{} & 0.122 [0.101, 0.141] & 0.596 [0.588, 0.604] \\
RADAR & \human{} vs. \llmtol{} & 0.646 [0.617, 0.670] & 0.903 [0.899, 0.908] \\
\bottomrule
\end{tabularx}
\end{table*}

\subsection{Robustness under \htol{} rewriting}
\label{subsec:rq2_h2l}

\begin{figure}[pos=t]
    \centering
    \includegraphics[width=0.92\textwidth]{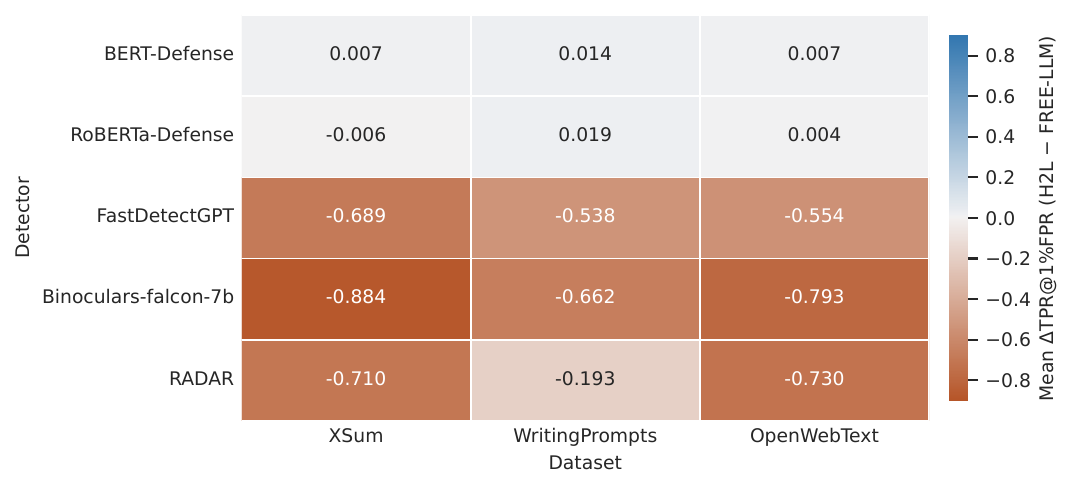}
    \caption{Mean $\Delta$\tprfpr{} under \htol{} by detector and dataset. Each cell is the mean across generators of the paired block-level difference between \htol{} and \freellm{} (\htol{} minus \freellm{}). Negative values indicate degradation relative to direct LLM generation.}
    \label{fig:delta_tpr_h2l}
\end{figure}

This analysis measures transfer from the \freellm{} baseline to the \htol{} regime. Table~\ref{tab:rq2_delta} reports paired target-minus-\freellm{} deltas for both rewriting regimes; this subsection focuses on the \htol{} columns. Negative values indicate degradation relative to direct LLM generation. Figure~\ref{fig:delta_tpr_h2l} shows how mean $\Delta$\tprfpr{} varies across detectors and datasets after averaging the paired block-level deltas across generators.

The strongest baseline detectors experience the largest \htol{} losses. FastDetectGPT decreases from \tprfpr{} $0.912$ in \freellm{} to $0.308$ in \htol{}, corresponding to $\Delta$\tprfpr{} $=-0.605$. Binoculars-falcon-7b decreases from $0.935$ to $0.151$, corresponding to $\Delta$\tprfpr{} $=-0.784$. RADAR decreases from $0.668$ to $0.122$, corresponding to $\Delta$\tprfpr{} $=-0.546$. Figure~\ref{fig:delta_tpr_h2l} shows that these losses are not uniform across datasets: Binoculars-falcon-7b and RADAR show particularly large degradation on XSum and OpenWebText, whereas WritingPrompts is comparatively less severe for some detectors.

The AUROC losses are also visible but less operationally severe than the low-FPR losses. FastDetectGPT drops by $-0.117$ AUROC, while Binoculars-falcon-7b and RADAR drop by $-0.316$ and $-0.318$, respectively. BERT-Defense and RoBERTa-Defense show small or positive deltas, but these values should not be interpreted as robustness: both detectors have near-zero \tprfpr{} in the baseline and remain near zero under \htol{}.

\begin{table*}[t]
\centering
\small
\caption{Regime deltas relative to \freellm{}. Deltas are computed as paired target-minus-\freellm{} differences within dataset $\times$ generator blocks and then macro-averaged. Negative values indicate degradation.}
\label{tab:rq2_delta}
\resizebox{\textwidth}{!}{\begin{tabular}{lcccc}
\toprule
Detector & $\Delta$\tprfpr{} \htol{} & $\Delta$AUROC \htol{} & $\Delta$\tprfpr{} \llmtol{} & $\Delta$AUROC \llmtol{} \\
\midrule
BERT-Defense & 0.010 [0.007, 0.014] & 0.243 [0.235, 0.251] & 0.001 [-0.001, 0.002] & 0.023 [0.020, 0.025] \\
RoBERTa-Defense & 0.006 [-0.002, 0.014] & 0.017 [0.008, 0.025] & -0.003 [-0.008, 0.001] & -0.032 [-0.036, -0.027] \\
FastDetectGPT & -0.605 [-0.645, -0.569] & -0.117 [-0.122, -0.111] & -0.130 [-0.146, -0.115] & -0.020 [-0.021, -0.018] \\
Binoculars-falcon-7b & -0.784 [-0.802, -0.767] & -0.316 [-0.325, -0.308] & -0.104 [-0.114, -0.096] & -0.031 [-0.034, -0.028] \\
RADAR & -0.546 [-0.575, -0.514] & -0.318 [-0.326, -0.310] & -0.023 [-0.030, -0.015] & -0.010 [-0.012, -0.008] \\
\bottomrule
\end{tabular}}
\end{table*}

Thus, \htol{} substantially weakens detector performance for detectors that are effective in the standard baseline condition. The effect is strongest at the low false-positive operating point.

\subsection{Robustness under same-generator \llmtol{} second pass}
\label{subsec:rq3_llm2l}

This analysis evaluates whether a second pass through the same generator weakens detector signals in already LLM-origin text. The \llmtol{} deltas in Table~\ref{tab:rq2_delta} are substantially smaller than the corresponding \htol{} deltas for detectors with strong baselines.

\begin{figure}[pos=ht!]
    \centering
    \includegraphics[width=0.92\textwidth]{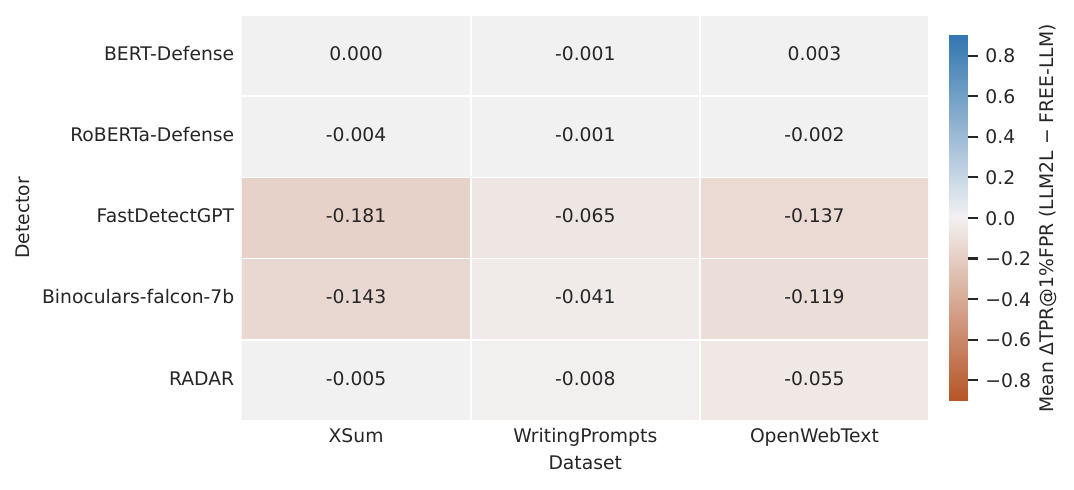}
    \caption{Mean $\Delta$\tprfpr{} under \llmtol{} by detector and dataset. Each cell is the mean across generators of the paired block-level difference between \llmtol{} and \freellm{} (\llmtol{} minus \freellm{}). Negative values indicate degradation relative to direct LLM generation.}
    \label{fig:delta_tpr_llm2l}
\end{figure}

Figure~\ref{fig:delta_tpr_llm2l} reports the mean $\Delta$\tprfpr{} under \llmtol{} for each detector and dataset after averaging the paired block-level deltas across generators.

FastDetectGPT decreases from \tprfpr{} $0.912$ in \freellm{} to $0.783$ in \llmtol{}, with $\Delta$\tprfpr{} $=-0.130$. Binoculars-falcon-7b decreases from $0.935$ to $0.830$, with $\Delta$\tprfpr{} $=-0.104$. RADAR is nearly stable, decreasing from $0.668$ to $0.646$, with $\Delta$\tprfpr{} $=-0.023$. The corresponding AUROC deltas are also small: $-0.020$ for FastDetectGPT, $-0.031$ for Binoculars-falcon-7b, and $-0.010$ for RADAR.

The dataset-level pattern in Figure~\ref{fig:delta_tpr_llm2l} confirms that \llmtol{} degradation is weaker than the \htol{} degradation reported in Figure~\ref{fig:delta_tpr_h2l}. Same-generator second-pass rewriting therefore does not erase machine-origin signals to the same degree as human-origin LLM rewriting. For detectors that are informative in the baseline benchmark, \llmtol{} remains much closer to \freellm{} than to \htol{}.

\subsection{Operational \htol{}--\llmtol{} gap under LLM-mediated surface}
\label{subsec:rq4_source_origin}

This analysis compares \htol{} and \llmtol{} directly. Both regimes have an LLM-mediated final surface, but they differ in source origin and in observed transformation strength: \htol{} starts from human-authored content and undergoes a stronger measured transformation, whereas \llmtol{} starts from LLM-origin content. Table~\ref{tab:source_origin_gap} reports the operational \llmtol{} minus \htol{} gap with bootstrap uncertainty. Gaps are computed as paired \llmtol{}-minus-\htol{} differences within dataset $\times$ generator blocks and macro-averaged across blocks.

\begin{figure}[pos=t]
    \centering
    \includegraphics[width=0.90\textwidth]{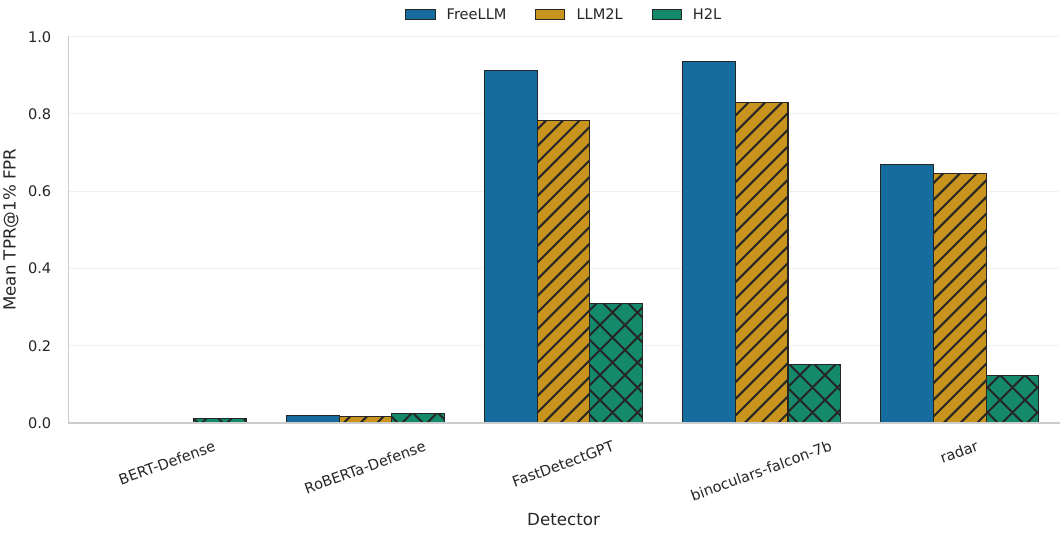}
    \caption{Detection performance across generation regimes. Bars show the detector-level mean \tprfpr{} for \freellm{}, \llmtol{}, and \htol{}, macro-averaged across dataset $\times$ generator blocks. Error bars denote block-structured bootstrap 95\% confidence intervals. The contrast between \htol{} and \llmtol{} visualizes the source-origin gap under LLM-mediated surface.}
    \label{fig:detection_gap}
\end{figure}

Figure~\ref{fig:detection_gap} provides the complementary absolute-performance view, comparing mean \tprfpr{} across \freellm{}, \llmtol{}, and \htol{} for every detector. It shows that the large negative \htol{} deltas in Figure~\ref{fig:delta_tpr_h2l} arise from a sharp reduction relative to the direct-generation baseline, whereas \llmtol{} generally remains much closer to \freellm{}.

For the strongest baseline detectors, \llmtol{} is much more detectable than \htol{}. Binoculars-falcon-7b has a \tprfpr{} gap of $0.680$ (95\% CI $[0.661, 0.698]$), RADAR has a gap of $0.524$ ($[0.493, 0.551]$), and FastDetectGPT has a gap of $0.475$ ($[0.444, 0.511]$). All three intervals exclude zero. These gaps show that the presence of an LLM-mediated final surface is not sufficient to explain detector performance: \htol{} and \llmtol{} remain substantially different at the low-FPR operating point.

However, the textual diagnostics in Section~\ref{subsec:textual_transformation} show that \htol{} also applies stronger surface transformation than \llmtol{} (NED $0.602$ vs. $0.348$; Jaccard $0.463$ vs. $0.587$). The gap between the two regimes therefore reflects some combination of source origin and transformation strength; the two factors are partially confounded in the current design. For this reason, we interpret the result as an operational source-origin-associated gap rather than as a causal estimate of source origin alone.

For BERT-Defense and RoBERTa-Defense, the operational gaps are not diagnostically useful because both detectors have near-zero \tprfpr{} in all regimes. Their small or negative gaps reflect poor operating-point performance rather than evidence of robustness.

\begin{table}[t]
\centering
\small
\caption{Paired operational gap between \llmtol{} and \htol{}. Positive values indicate that \llmtol{} is more detectable than \htol{}. Values are macro-averaged across dataset $\times$ generator blocks; brackets report block-structured bootstrap 95\% confidence intervals.}
\label{tab:source_origin_gap}
\begin{tabular*}{\tblwidth}{@{} Lcc@{}}
\toprule
Detector & \tprfpr{} gap [95\% CI] & AUROC gap [95\% CI] \\
\midrule
BERT-Defense & -0.010 [-0.013, -0.007] & -0.220 [-0.228, -0.213] \\
RoBERTa-Defense & -0.009 [-0.017, -0.002] & -0.048 [-0.057, -0.040] \\
FastDetectGPT & 0.475 [0.444, 0.511] & 0.097 [0.092, 0.102] \\
Binoculars-falcon-7b & 0.680 [0.661, 0.698] & 0.285 [0.277, 0.294] \\
RADAR & 0.524 [0.493, 0.551] & 0.308 [0.300, 0.315] \\
\bottomrule
\end{tabular*}
\end{table}

\subsection{Detector-family, dataset, and generator heterogeneity}
\label{subsec:rq5_heterogeneity}

This analysis examines whether the \htol{} degradation is uniform across detectors, datasets, and generator models. The full AUROC and \tprfpr{} heatmaps for the \htol{} regime are reported in Appendix~\ref{app:h2l_heatmaps}; each panel corresponds to one detector, and each cell represents a generator $\times$ dataset block.

The largest differences are observed across detector families. FastDetectGPT is the most robust detector under \htol{}, although it still loses substantial \tprfpr{}. Binoculars-falcon-7b and RADAR are strong in \freellm{} and \llmtol{} but degrade sharply in \htol{}. BERT-Defense and RoBERTa-Defense remain weak at the low-FPR operating point across regimes.

Dataset effects are also visible. Under \htol{}, Binoculars-falcon-7b is comparatively stronger on WritingPrompts but degrades on OpenWebText and XSum. RADAR shows a different pattern: it can obtain strong AUROC on XSum while retaining only moderate \tprfpr{}, illustrating that ranking quality and conservative-threshold recall can diverge.

Generator-level variation is present but less uniform than dataset-level variation. No generator is uniformly easy or difficult across all detectors. However, qwen25\_7b appears more challenging for selected detectors, especially in low-FPR recall, while llama32\_3b is often easier to detect for FastDetectGPT and Binoculars-falcon-7b. These observations should be interpreted as block-level heterogeneity rather than as causal properties of individual generators.

Overall, this heterogeneity analysis shows that robustness is driven primarily by the detector family, with additional variation in datasets and generators that would be hidden by pooled benchmark-level averages.
\subsection{Textual transformation analysis}
\label{subsec:textual_transformation}

As a supporting diagnostic analysis, we compare source and target texts using target-to-source word ratio, token-level normalized edit distance, lexical overlap, and semantic similarity. For \htol{}, the source is the original \human{} text and the target is its LLM-mediated rewrite. For \llmtol{}, the source is the corresponding \freellm{} output and the target is the same-generator second-pass output.

\begin{figure}[pos=ht!]
    \centering
    \includegraphics[width=0.9\linewidth]{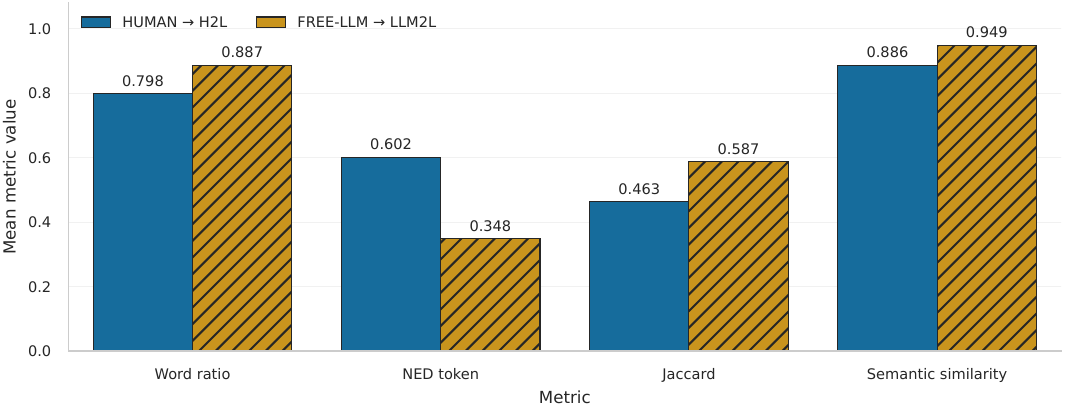}
    \captionof{figure}{Textual change induced by the two rewriting paths. \htol{} corresponds to \human{} $\rightarrow$ \htol{}, whereas \llmtol{} corresponds to \freellm{} $\rightarrow$ \llmtol{}. The figure shows that \htol{} introduces stronger surface changes, while \llmtol{} is a more conservative second-pass rewrite. Error bars denote block-structured bootstrap 95\% confidence intervals.}
    \label{fig:textual_rewriting_paths}
\end{figure}

Table~\ref{tab:rewriting-paths} compares the textual change induced by the two rewriting paths, while Figure~\ref{fig:textual_rewriting_paths} provides a compact visual summary. \htol{} produces shorter outputs than \llmtol{}, with a word ratio of $0.798$ compared with $0.887$. It also introduces stronger token-level changes, with normalized edit distance $0.602$ compared with $0.348$ for \llmtol{}. Lexical overlap is lower in \htol{} ($0.463$) than in \llmtol{} ($0.587$). Both paths retain high semantic similarity, although \llmtol{} is more conservative: semantic similarity is $0.886$ for \htol{} and $0.949$ for \llmtol{}.

These diagnostics characterize \htol{} as a broadly meaning-preserving but more surface-altering rewrite than \llmtol{}. This pattern is consistent with the larger detection loss under \htol{}, but it does not establish that any single textual feature causes the loss.

\begin{table*}[t]
\centering
\small
\caption{Textual change induced by the two rewriting paths. \htol{} measures the transformation from \human{} to \htol{}, whereas \llmtol{} measures the transformation from \freellm{} to \llmtol{}. Values report mean estimates with 95\% confidence intervals.}
\label{tab:rewriting-paths}
\begin{tabularx}{\textwidth}{lccX}
\toprule
Aspect & \human{} $\rightarrow$ \htol{} & \freellm{} $\rightarrow$ \llmtol{} & Interpretation \\
\midrule
Word ratio & 0.798 [0.746, 0.845] & 0.887 [0.857, 0.916] & More compression in \htol{} \\
Token NED & 0.602 [0.575, 0.628] & 0.348 [0.301, 0.395] & Greater surface change in \htol{} \\
Jaccard overlap & 0.463 [0.436, 0.489] & 0.587 [0.534, 0.648] & Less vocabulary overlap in \htol{} \\
Semantic similarity & 0.886 [0.866, 0.904] & 0.949 [0.939, 0.958] & \llmtol{} is more conservative \\
\bottomrule
\end{tabularx}
\end{table*}

\section{Discussion}
\label{sec:discussion}
The results have implications for detector robustness, benchmark validity, and deployment under LLM-mediated rewriting. Section~\ref{subsec:implications} makes explicit how these results advance theory relative to prior paraphrase-robustness and benchmark-transfer work, and what they imply for detector selection and deployment in practice.

\subsection{Interpretation}
\label{subsec:main_empirical_finding}
The main empirical finding is a benchmark-transfer failure: standard \human{} vs. \freellm{} evaluation overestimates robustness when the target includes human-origin LLM-mediated rewriting. FastDetectGPT, Binoculars-falcon-7b, and RADAR remain comparatively stable under same-generator \llmtol{} but degrade sharply under \htol{}, so an additional LLM pass does not by itself make LLM-origin text resemble the harder \htol{} condition. This is a benchmark-transfer failure across matched regimes, however, not a pure causal effect of source origin.\label{subsec:identification_scope} \htol{} also changes its source more extensively than \llmtol{} (Section~\ref{subsec:textual_transformation}), so the \htol{}--\llmtol{} gap should be read as an operational source-origin-associated gap under LLM-mediated surface, rather than as evidence that source origin alone causes degradation; isolating source origin would require rewriting conditions explicitly matched on transformation strength (see Section~\ref{sec:limitations} for how this bears on interpretation).

Two independent detection-method literatures corroborate the direction of this asymmetry through an entirely different mechanism, rewrite-induced similarity or edit distance rather than classifier scoring. \citet{huang-etal-2025-magret} show that a model's own rewrite of its output stays measurably closer to that model's typical generations than any human text, stable across generator families, decoding temperatures, and top-$p$ values. RAIDAR~\citep{mao2024raidar} and its generalization L2R~\citep{hao-etal-2025-learning} build on the complementary premise, that human-origin text is edited more than LLM-origin text under LLM rewriting, to construct a detector directly from that edit distance, validating the asymmetry across 21 domains and four generator families. Where we treat this asymmetry as a source of benchmark-transfer failure for existing classifier-based detectors, these works treat it as an exploitable detection signal in its own right; together, the two perspectives suggest that content-origin sensitivity under LLM-mediated rewriting is a structural property of current LLMs rather than an artifact of any one detector or dataset.

\subsection{Comparison with prior benchmarks}
\label{subsec:relation_prior_paraphrase}
Prior work shows that detector performance is sensitive to paraphrasing, rewriting, adversarial prompting, and benchmark construction~\citep{krishna2023paraphrasing,sadasivan2023reliably,shi-etal-2024-red,dugan2024raid}. PADBen studies paraphrase attack trajectories for human- and LLM-authored content, while HLPC measures the effects of human and LLM paraphrases at a 1\% FPR operating point~\citep{zha2025padben,lau2025effects}. \arb{} complements these studies by treating the problem as benchmark transfer: whether performance under conventional direct generation predicts performance across matched LLM-mediated authorship--surface regimes under a shared human reference distribution. The same-generator \llmtol{} control makes this a matched comparison of two LLM-mediated regimes within each dataset--generator block, revealing an asymmetry consistent with PADBen's distinction between authorship obfuscation and plagiarism evasion while avoiding a pooled ``rewritten'' class that would conceal differences between human- and LLM-origin inputs~\citep{zha2025padben}; the contribution is therefore not another demonstration that rewriting reduces detection performance, but evidence that direct-generation benchmarks transfer differently to two distinct rewriting paths.

The content-origin/linguistic-surface distinction underlying Table~\ref{tab:regimes} echoes the genesis-based notion taxonomy of \citet{dycke2026yourai}, who likewise separate what a text's tokens objectively are from the normative target a detector is asked to recognize, and who recommend that notion parameters, such as the minimum AI-token ratio required to call a document AI-generated, be stated explicitly rather than left implicit in the data-generation procedure. Our four-regime design can be read as instantiating their document-level notion twice, once with a human genesis and once with an LLM genesis, specifically to isolate the transfer gap between the two; our regime labels play the same explicitness role for the content-origin/surface-mediation split as their $\tau$ plays for genesis granularity.

A related asymmetry appears in \citet{baidya2026detectingmachinecomprehensivebenchmark}'s humanization study: rewriting already-LLM-generated text with a separate instruction-tuned model leaves AUROC unchanged or higher at light intensity for every detector, and even their heaviest setting keeps every detector above AUROC $0.857$, consistent with our own \llmtol{} results, where an LLM-mediated second pass over LLM-origin text produces only a small $\Delta$\tprfpr{} relative to \freellm{} (Section~\ref{subsec:rq3_llm2l}). Because their rewriting is applied exclusively to LLM-origin text, their result cannot indicate whether the same pipeline applied to human-origin text would reproduce the much sharper \htol{}-style degradation we observe; \arb{}'s matched design is what makes that comparison possible.

\subsection{Theoretical and Practical Implications}
\label{subsec:implications}

\textbf{Theoretical implications.} The core theoretical contribution is reframing paraphrase robustness as a benchmark-transfer problem rather than a single robustness scalar. Prior paraphrase-robustness studies typically report one degradation curve per detector under an increasingly aggressive rewriter~\citep{krishna2023paraphrasing,sadasivan2023reliably,shi-etal-2024-red}, which conflates two logically independent quantities: how much a rewriting step changes the surface form, and whether the content being rewritten originated from a human or a machine. \arb{}'s matched four-regime design separates these quantities by holding the rewriting instruction and generator fixed while varying only the origin of the input text (\htol{} vs. \llmtol{}). This isolates a previously under-specified property of current detectors and LLMs: detector scores are sensitive to content origin under LLM-mediated surface, not only to surface distance from the training distribution of direct machine text. This complements the rewrite-similarity and edit-distance mechanisms proposed by \citet{huang-etal-2025-magret}, \citet{mao2024raidar}, and \citet{hao-etal-2025-learning} (Section~\ref{subsec:main_empirical_finding}) with a classifier-level demonstration of the same asymmetry, suggesting that content-origin sensitivity is a structural property of the human-LLM rewriting relationship rather than an artifact specific to one detection paradigm (statistical, watermark-based, or edit-distance-based). It also refines the genesis-based notion framework of \citet{dycke2026yourai}: our results show empirically, rather than only conceptually, why a fixed genesis label and a fixed detection notion need to be crossed explicitly, since the same nominal ``LLM-involved'' text can sit far apart in detectability depending on which side of the human/LLM boundary supplied the original content.

\textbf{Practical implications.} These findings have direct consequences for how detectors are selected, evaluated, and deployed. First, procurement and audit decisions that rely on \human{} vs. \freellm{} benchmarks alone (the current de facto standard reported by most detector papers) risk substantially overestimating robustness for the increasingly common case of human drafts revised by an LLM, a workflow now standard in academic, journalistic, and professional writing. A detector that looks strong under direct-generation testing, such as FastDetectGPT or Binoculars-falcon-7b, can lose most of its low-FPR recall under \htol{} without any change in its published AUROC on the vendor's own benchmark. Second, the \htol{}--\llmtol{} gap gives practitioners a concrete pre-deployment test: before trusting a detector in a setting where human-authored, LLM-polished text is in scope (e.g., plagiarism review, academic integrity, or content-provenance pipelines), the detector should be evaluated on an \htol{}-style condition specifically, not inferred from \freellm{} or \llmtol{} results. Third, because degradation is uneven across detector families (RADAR and Binoculars degrade more under \htol{} than under adversarial \llmtol{}-style rewriting; BERT/RoBERTa baselines are weak throughout), a single aggregate leaderboard score is an unreliable basis for tool selection; deployers need regime-, domain-, and generator-stratified numbers of the kind \arb{} reports. We detail the operating-point and reporting practices that follow from this in Section~\ref{subsec:low_fpr_discussion}.

\subsection{Validity and deployment implications}
\label{subsec:low_fpr_discussion}
AUROC and \tprfpr{} capture different aspects of robustness. A detector may retain some global ranking ability while recovering few positives at a threshold constrained to misclassify only 1\% of human texts, a distinction that matters where false accusations carry fairness costs, including documented bias against non-native English writers~\citep{LIANG2023100779}, and that supports recent calls to assess low-FPR recall or threshold stability rather than treat detector scores as conclusive evidence~\citep{website-new-ai-classifier-for-indicating-ai-written-text,lau2025effects,ayoobi2025shield,chen2025divscore,masrour2025damage}.

At this operating point, detector families separate clearly, and not always in the direction a paraphrase-robustness label would suggest. RADAR, although designed for paraphrase robustness~\citep{hu2023radar}, remains reliable under \llmtol{} but degrades under \htol{}; this stability under \llmtol{} should be read against its attacker model, since our \freellm{}$\to$\llmtol{} rewrite is a plain, meaning-preserving instruction with no detector-evasion objective, whereas a detector-in-the-loop attack that explicitly selects paraphrase candidates to minimize a target detector's score drives RADAR's accuracy from 90.0\% to 45.4\% on the same class of LLM-generated text~\citep{huang-etal-2025-tempparaphraser}. The two results isolate different factors, content origin in our matched design and an explicit adversarial objective in theirs, and together suggest that RADAR's paraphrase robustness holds for non-adversarial LLM-mediated rewriting but not for rewriting optimized against it. The supervised BERT/RoBERTa baselines provide little utility across regimes, consistent with concerns about distribution shift and benchmark transfer~\citep{10179387,li2024mage,Schaaff2024} and with a broader pattern in which the same backbones reach near-ceiling in-domain scores yet degrade sharply once the evaluation distribution shifts~\citep{mady2026featureaugmentedtransformersrobustaitext}; here the shift is content origin under a fixed low-FPR bar, which BERT-Defense and RoBERTa-Defense fail to clear even in the \freellm{} baseline.

Robustness cannot be reduced to detector family alone: the block-level analyses show substantial dataset effects and less uniform generator effects, in line with evidence that rankings change with domain, generator, task, and metric~\citep{dugan2024raid,wu2024detectrl,prohl2024benchmarking,stowe2026spotlights,baidya2026detectingmachinecomprehensivebenchmark,mady2026featureaugmentedtransformersrobustaitext}, so aggregate results should be accompanied by stratified diagnostics rather than interpreted as regime- or domain-invariant properties.

These patterns motivate concrete evaluation and deployment practice. Benchmarks should define the positive class explicitly, reporting \freellm{}, \htol{}, and \llmtol{} separately rather than collapsing them into one LLM-involved class, and should preserve the matched structure of the data: compute regime deltas within dataset $\times$ generator blocks, distinguish macro-averages from pooled estimates, report uncertainty that respects the hierarchy, and retain domain- and generator-level diagnostics, aligning with emerging benchmarks centered on robustness, domain shift, and mixed authorship~\citep{wang2024m4gtbench,ayoobi2025shield,zha2025padben}. Holding \tprfpr{} at a single, fixed 1\% threshold across all regimes, datasets, and generators mirrors an independently converging recommendation to calibrate a decision threshold once on held-out data and keep it fixed rather than re-tuned per target distribution, since re-tuning at test time can otherwise mask the operating-point trade-offs that matter for deployment~\citep{mady2026featureaugmentedtransformersrobustaitext}. The same distinctions apply in deployment: detector scores are not regime-invariant evidence of machine authorship, so validation data should reflect the intended use case and include human-origin revised text whenever it is in scope, and reports should present performance at a prespecified false-positive constraint, false-positive behavior on relevant human populations, domain-specific estimates, and uncertainty. Given the residual error and distribution sensitivity observed here, detector output is better treated as one uncertain signal than as a stand-alone basis for high-stakes authorship judgments.

\subsection{Threats to Validity and Limitations}
\label{sec:limitations}

\textbf{Construct validity.} The benchmark evaluates score separability between \human{} texts and texts produced or mediated by LLMs. It does not establish that a detector identifies a single construct such as authorship, intent, originality, plagiarism, or amount of AI assistance. We mitigate this risk by defining four explicit regimes in terms of content origin and linguistic surface, and by interpreting detector scores as regime-specific separability estimates rather than as direct authorship judgments. This distinction is especially important for \htol{}, where the content is human-origin but the final linguistic surface is LLM-mediated.

\textbf{Internal validity.} Prompt wording, decoding settings, source preprocessing, and sampling can affect the final text distribution. We mitigate these risks through fixed prompts, fixed decoding settings, matched source items, generator-specific blocks, seeded length-stratified sampling, and a single preprocessing pipeline applied before sampling. Nevertheless, different rewriting prompts, temperatures, decoding strategies, or preprocessing choices could produce different surface properties. Residual formatting artifacts or source-specific cues cannot be ruled out completely. The same-generator \llmtol{} condition controls the second pass within each generator family, but it does not cover cross-model rewriting, multi-step rewriting, or human-edited rewriting. Because no post-hoc semantic or length filtering was applied to the released generated texts (Section~\ref{subsec:dataset_description}), a small union of 0.67\% of generated texts are atypically short, refusal-like, or non-English; at the sample sizes used for block-level and macro-averaged estimation, this is expected to contribute negligible additional noise, but it remains a residual source of measurement error at the level of individual scored items.

\textbf{Transformation-strength confounding.} A central limitation is that \htol{} induces stronger surface change than same-generator \llmtol{} (Token NED $0.602$ vs. $0.348$; Jaccard overlap $0.463$ vs. $0.587$). We mitigate this issue by reporting textual transformation diagnostics and by avoiding a causal claim that source origin alone explains the \htol{}--\llmtol{} gap. The current design supports a source-origin-associated interpretation, but it does not fully disentangle source origin from transformation intensity. A stronger design would include same-intensity rewriting conditions or additional controls that explicitly match transformation strength across \htol{} and \llmtol{}.

\textbf{Implementation validity.} Detectors were evaluated using released implementations or paper-recommended configurations whenever available, with hardware-driven adjustments only for execution feasibility. We mitigate implementation bias by keeping each detector configuration fixed across all regimes, datasets, and generator models, so that detector retuning does not confound comparisons among \freellm{}, \htol{}, and \llmtol{}. However, implementation details such as maximum observed sequence length, numerical precision, quantization, tokenizer behavior, and library versions can affect detector scores. The reported results should therefore be interpreted as estimates for the evaluated implementations, not as universal properties of the underlying detector families.

\textbf{External validity.} The benchmark uses English texts from XSum, WritingPrompts, and OpenWebText; four open-weight generator families; and the detector implementations listed in Table~\ref{tab:detectors}. We mitigate over-specialization by using multiple domains, multiple generator families, and complementary detector families. However, the results should not be generalized without further evaluation to other languages, longer documents, specialized professional domains, closed-source generators, multimodal content, or detectors trained specifically on \htol{} examples. Bilingual evidence from CUDRT indicates that language and operation type are both first-order factors for detector generalization~\citep{10.1145/3779427}, so the English-only scope here is a substantive limitation rather than a minor one. The dataset choices provide domain diversity, but they are not exhaustive.

\textbf{Statistical conclusion validity.} The data are matched at the dataset $\times$ generator block level, and regime comparisons share source material within each block. Treating pooled texts as independent would understate uncertainty and obscure the paired design. We mitigate this risk by using block-level estimates, paired deltas within blocks, macro-averages across blocks, and a block-structured bootstrap with 5,000 resamples for confidence intervals. These intervals quantify uncertainty in aggregate estimates, but they do not replace block-level heterogeneity analysis. Because multiple detector-regime comparisons are reported, results should be interpreted as benchmark estimates rather than as isolated null-hypothesis tests.

\textbf{Benchmark realism and deployment representativeness.} \htol{} and \llmtol{} approximate controlled rewriting workflows, but real users may interact with LLMs iteratively, manually edit outputs, combine multiple models, use different instructions, or mix generated and human-written passages within a single document. We mitigate this limitation by operationally varying authorship and surface regimes in a matched design, which is more realistic than direct-generation-only evaluation. However, the benchmark remains a controlled approximation of human--LLM writing workflows rather than an observational study of real-world writing behavior.
\subsection{Ethical Considerations}
\label{sec:ethics}

This work is intended to improve the evaluation of AI-text detectors and to reduce overconfident deployment under distribution shift. The benchmark involves generation and rewriting procedures that could also be interpreted as evasion-relevant. We mitigate this dual-use risk by framing rewriting as an evaluation condition, reporting detector-side implications, and avoiding operational guidance for bypassing deployed systems. \arb{}'s scope is therefore narrower than dedicated evasion research: unlike gradient-based evaders explicitly optimized against a victim detector and demonstrated against deployed commercial systems~\citep{meng2025gradescape}, our \htol{} and \llmtol{} rewrites use a single fixed, non-adversarial instruction with no detector in the loop, and we do not target or report evasion rates against any specific product. The study uses existing public datasets and locally generated text variants; no human subjects or user studies are involved. The released \arb{} assets are intended for scientific benchmarking, auditing, and robustness evaluation rather than for high-stakes authorship accusations or operational bypassing of deployed systems. Dataset licenses and provenance are documented in the released artifacts.

\section{Conclusion}
\label{sec:conclusion}

This paper presented a matched quantitative benchmark of AI-text detectors across authorship and rewriting regimes. \arb{} operationally contrasts \human{}, \freellm{}, \htol{}, and same-generator \llmtol{} under a benchmark-transfer design. The study evaluates existing detectors as objects of benchmarking and reports \tprfpr{}, AUROC, paired deltas, and operational \htol{}--\llmtol{} gaps over dataset $\times$ generator blocks.

The results show that performance estimated under the standard \human{} vs. \freellm{} condition does not necessarily transfer to \htol{}. FastDetectGPT, Binoculars-falcon-7b, and RADAR perform well on direct LLM generation and remain substantially closer to that baseline under \llmtol{}, but lose much more low-FPR recall under \htol{}. BERT-Defense and RoBERTa-Defense remain weak at \tprfpr{} across regimes. The comparison between \htol{} and \llmtol{} suggests that rewriting alone does not fully explain the degradation; source origin and transformation strength both plausibly contribute to the gap (Section~\ref{subsec:identification_scope}).

These findings support detector evaluations that include human-origin rewriting, same-generator second-pass controls, block-level paired deltas, domain and generator breakdowns, and low false-positive operating points in addition to AUROC. This recommendation is consistent with recent benchmark work emphasizing mixed-authorship text, humanization, detector stability, and metric sensitivity~\citep{wang2024m4gtbench,masrour2025damage,ayoobi2025shield,stowe2026spotlights}. The results also support a more precise deployment vocabulary: direct machine generation and LLM-mediated rewriting are not equivalent detection targets.

Future work should extend \arb{} along three main directions. First, stronger causal isolation of source origin requires same-intensity rewriting controls, in which \htol{} and \llmtol{} are matched not only by the source item and the generator block but also by the transformation strength. Second, broader external validation should include additional languages, longer documents, specialized professional domains, closed-source generators, cross-model rewriting, and mixed-authorship documents with paragraph- or sentence-level attribution. Third, future detector evaluations should test whether detectors trained or calibrated on \htol{} examples generalize to unseen rewriting styles, domains, and generators, rather than only improving on the specific benchmark distribution.

\section*{Declaration of Generative AI and AI-assisted Technologies in the Writing Process}

During the preparation of this work, the authors used generative AI and AI-assisted writing tools to support language editing, improve readability, and refine the manuscript's academic presentation. These tools were used for wording, grammar, stylistic revision, and organizing explanatory text.

Generative AI tools were also used to assist with non-substantive drafting support, such as improving section transitions, clarifying methodological descriptions, and formatting parts of the manuscript. They were not used to generate the experimental results, detector scores, statistical estimates, tables, or figures reported in the study. The benchmark construction, detector evaluation, metric computation, bootstrap analysis, and interpretation of results were conducted and verified by the authors.

All AI-assisted content was reviewed, edited, and validated by the authors to ensure accuracy, consistency with the experimental evidence, and alignment with the paper's claims. The authors take full responsibility for the content of the manuscript.

\section*{Data Availability}
\label{sec:data_availability}

The text dataset underlying \arb{} is publicly released on Hugging Face~\footnote{The url will be inserted after the anonymized review process} 

The public release contains all four regimes used in the benchmark (\human{}, \freellm{}, \htol{}, and \llmtol{}) and is distributed as a Hugging Face-compatible Parquet dataset that can be loaded directly with the \texttt{datasets} library. Each released row corresponds to one text sample and includes stable identifiers, the text, regime labels, content-origin and surface-origin labels, source-dataset provenance, generator-model metadata, source and pairing indices, a normalized-text SHA-256 hash, and word-count metadata. The release exposes the fields \texttt{id}, \texttt{text}, \texttt{label}, \texttt{label\_id}, \texttt{regime}, \texttt{source\_dataset}, \texttt{source\_dataset\_short}, \texttt{generator\_model}, \texttt{source\_index}, \texttt{pair\_id}, \texttt{text\_sha256}, and \texttt{word\_count}.

The released \arb{} text collection is derived from XSum, WritingPrompts, and OpenWebText, and includes rewrites produced by Gemma 2 9B, Llama 3.2 3B, Mistral 7B, and Qwen2.5 7B. The source code for data processing, evaluation, and figure/table reproduction is released separately in the Git repository (\url{https://anonymous.4open.science/r/arb-0E1C/}). The public dataset and repository jointly provide the text samples, pairing metadata, detector scores, block-level metrics, configuration files, and reproduction scripts required to reproduce the analyses reported in this paper. 
The configuration files are fixed using the \texttt{Hydra} Python configuration framework~\citep{Yadan2019Hydra}. The framework ensures reproducible, explicit settings for the dataset, generator, detector, decoding, and bootstrap. 

The dataset is released under the Apache License 2.0, with the caveat that users remain responsible for respecting the licenses and usage terms of the original source datasets (XSum, WritingPrompts, and OpenWebText; licensing details for each are reported in Section~\ref{subsec:dataset_description}).

\bibliographystyle{cas-model2-names-apa-year}

\appendix

\section{Prompt Templates}
\label{app:prompts}
The following appendix describes the prompt templates used to generate the ARB-Dataset texts. Each prompt contains specific parameters: \freellm{} \texttt{\{topic\}} depends on the source dataset, while \texttt{\{source\_text\}} and \texttt{\{free\_llm\_text\}} are, respectively, the human text of the source dataset and the LLM-generated text produced using the \freellm{} prompt (see Figure~\ref{fig:arb-construction-pipeline}). 

The design details of the prompts are reported in Section~\ref{subsec:prompt_protocol}.

\subsection{System prompt}
\begin{quote}\small
You are a text rewriting and generation engine for a scientific benchmark. Return only the requested text. Do not add explanations, comments, markdown, headings, or prefaces.
\end{quote}

\subsection{\freellm{} prompt}
\begin{quote}\small
Write a fluent, self-contained English text about the following topic.

Constraints:
Use your own wording and structure. Do not refer to the existence of a source text. Do not include headings or bullet points. Keep the length between \{min\_words\} and \{max\_words\} words. Return only the generated text.

Topic: \{topic\}
\end{quote}

\subsection{\htol{} prompt}
\begin{quote}\small
Rewrite the following text in fluent natural English.

Constraints:
Preserve the original meaning. Preserve factual claims, entities, and relationships. Do not add new information. Do not remove important information. Change wording and sentence structure where possible. Keep approximately the same length. Return only the rewritten text.

Text: \{source\_text\}
\end{quote}

\subsection{\llmtol{} prompt}
\begin{quote}\small
Rewrite the following text in fluent natural English.

Constraints:
Preserve the original meaning. Do not add new information. Do not remove important information. Change wording and sentence structure where possible. Keep approximately the same length. Return only the rewritten text.

Text: \{free\_llm\_text\}
\end{quote}
\newpage
\section{\htol{} block-level heterogeneity heatmaps}
\label{app:h2l_heatmaps}

\begin{figure}[pos=ht!]
    \centering
    \includegraphics[width=0.8\linewidth]{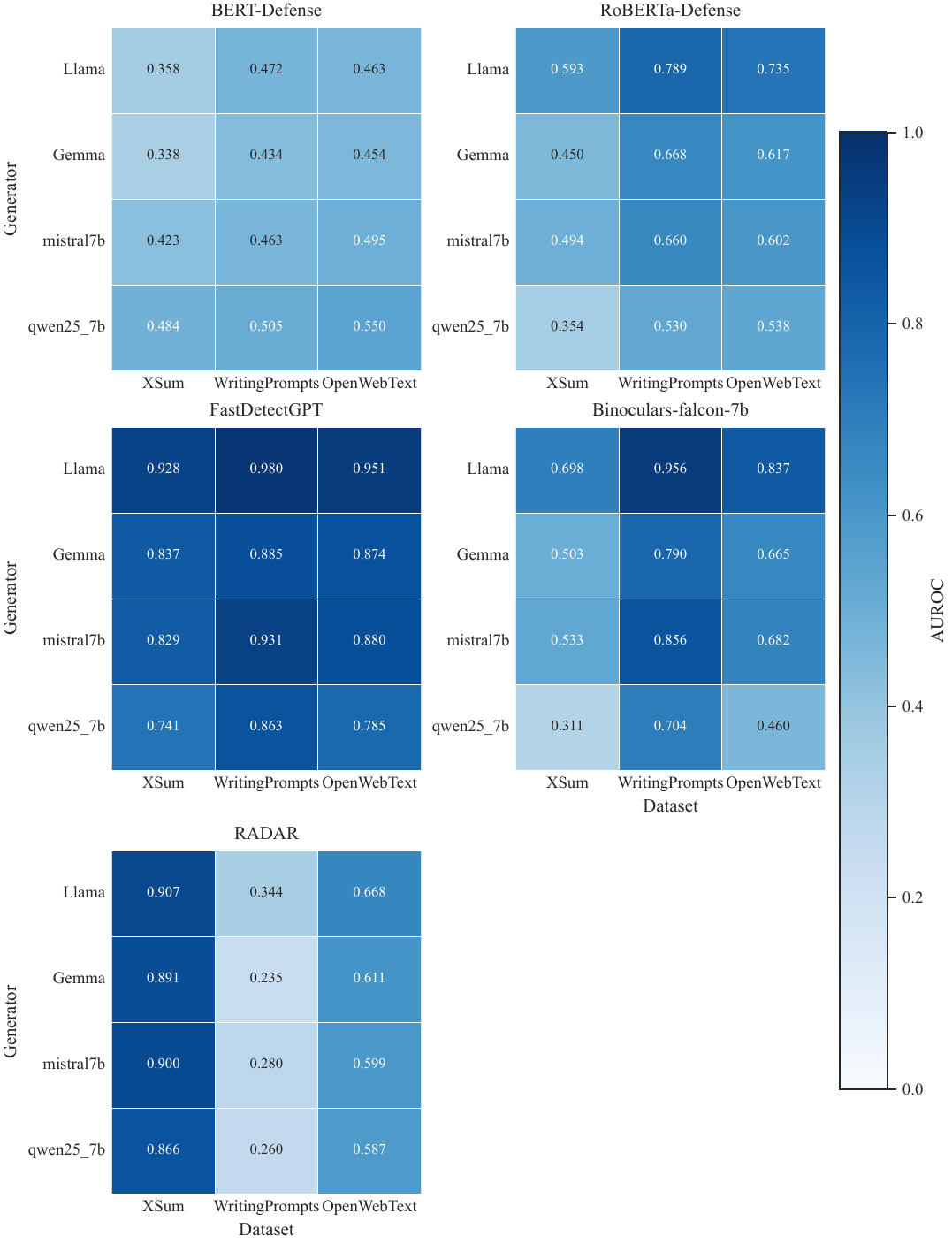}
    \caption{AUROC heatmaps for the \htol{} regime. Each panel corresponds to one detector; rows are generator models, and columns are datasets. Values are averaged within each generator $\times$ dataset cell. All panels share the same 0--1 color scale.}
    \label{fig:heatmap_h2l_auroc}
\end{figure}

\begin{figure}[pos=ht!]
    \centering
    \includegraphics[width=0.8\linewidth]{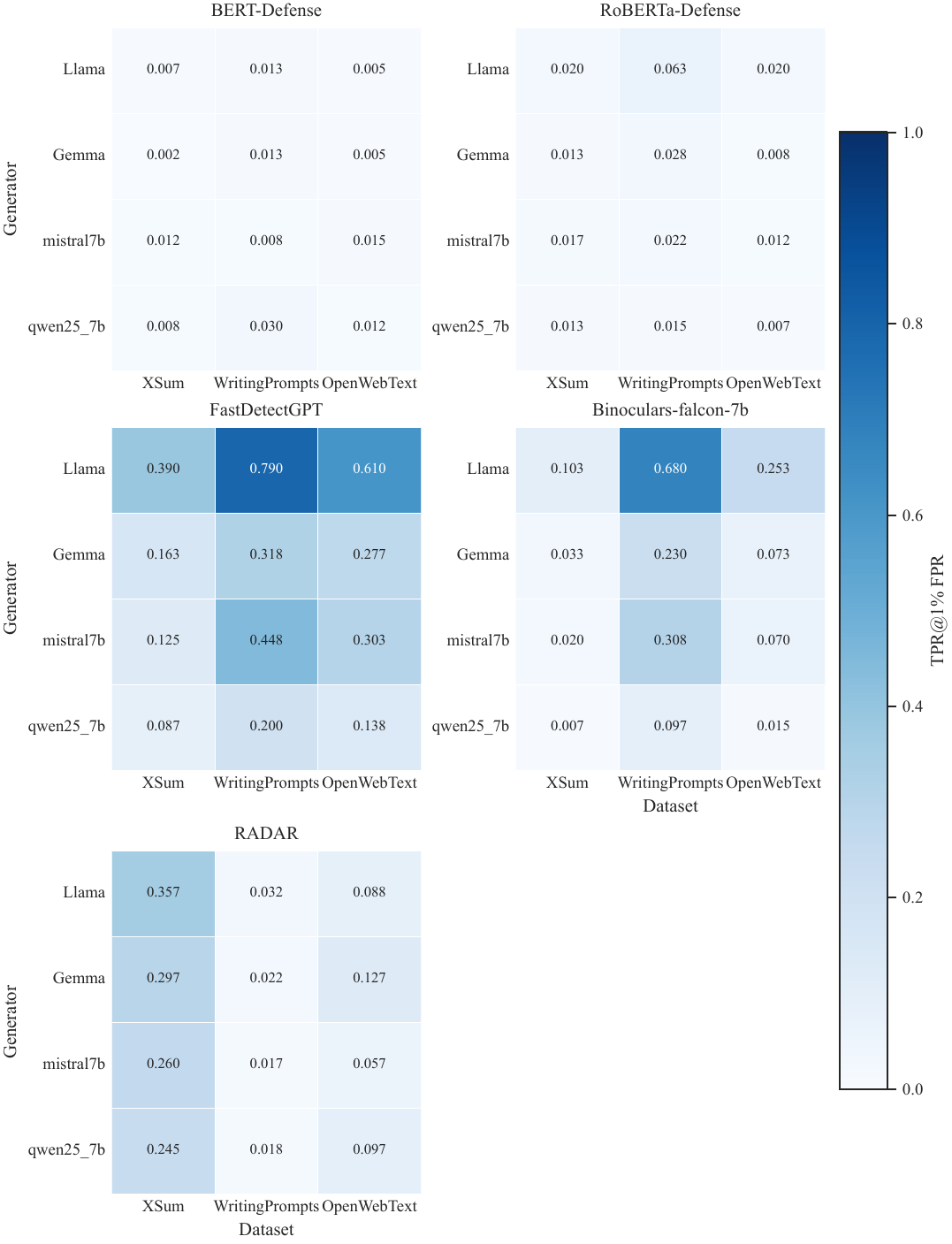}
    \caption{\tprfpr{} heatmaps for the \htol{} regime. Each panel corresponds to one detector; rows are generator models and columns are datasets. Values are averaged within each generator $\times$ dataset cell. All panels share the same 0--1 color scale.}
    \label{fig:heatmap_h2l_tpr}
\end{figure}

Figures~\ref{fig:heatmap_h2l_auroc} and~\ref{fig:heatmap_h2l_tpr} report the full block-level heterogeneity analysis for the \htol{} regime. Each panel corresponds to one detector, and each cell represents a generator $\times$ dataset block.

 \end{document}